\pdfoutput=1
\documentclass[11pt]{article}
\usepackage{amsmath}
\usepackage{amsfonts}
\usepackage{rotating}

\usepackage{enumitem}
\usepackage{multirow}
\usepackage{tcolorbox}
\tcbuselibrary{listingsutf8}
\usepackage{multicol}

\usepackage{amsmath,amssymb}
\usepackage[linesnumbered,ruled,vlined]{algorithm2e}

\SetKwProg{Fn}{Function}{:}{}  % \Fn{Header}{...}
\SetKwFunction{Parse}{ParseTime}
\SetKwFunction{Rank}{PrecRank}
\SetKw{Return}{return}         % so \Return{...} works

\usepackage{acl}
\usepackage{booktabs}
\usepackage{times}
\usepackage{latexsym}

\usepackage[T1]{fontenc}
\usepackage[utf8]{inputenc}

\usepackage{microtype}

\usepackage{inconsolata}

\usepackage{graphicx}

\title{RASTeR: Robust, Agentic, and Structured Temporal Reasoning}

\author{
\textbf{Dan Schumacher}\textsuperscript{$\dagger$},
\textbf{Fatemeh Haji}\textsuperscript{$\dagger$},
\textbf{Tara Grey}\textsuperscript{$\dagger$},
\textbf{Niharika Bandlamudi}\textsuperscript{$\dagger$},
\textbf{Nupoor Karnik}\textsuperscript{$\dagger$} \\
\textbf{Gagana Uday Kumar}\textsuperscript{$\dagger$},
\textbf{Cho-Yu Jason Chiang}\textsuperscript{$\S$},
\textbf{Peyman Najafirad}\textsuperscript{$\dagger$},\\
\textbf{Nishant Vishwamitra}\textsuperscript{$\dagger$}, and
\textbf{Anthony Rios}\textsuperscript{$\dagger$}
\\
\textsuperscript{$\dagger$}\,University of Texas at San Antonio,
\textsuperscript{$\S$}\,Peraton Labs
\\
\texttt{\{daniel.schumacher, anthony.rios\}@utsa.edu}
}

\begin{document}
\maketitle

\begin{abstract}
Temporal question answering (TQA) remains a challenge for large language models (LLMs), particularly when retrieved content may be irrelevant, outdated, or temporally inconsistent. This is especially critical in applications like clinical event ordering, and policy tracking, which require reliable temporal reasoning even under noisy or outdated information. To address this challenge, we introduce RASTeR: \textbf{R}obust, \textbf{A}gentic, and \textbf{S}tructured, \textbf{Te}mporal \textbf{R}easoning, a prompting framework that separates context evaluation from answer generation. RASTeR first assesses the relevance and temporal coherence of the retrieved context, then constructs a temporal knolwedge graph (TKG) to better facilitate reasoning. When inconsistencies are detected, RASTeR selectively corrects or discards context before generating an answer. Across multiple datasets and LLMs, RASTeR consistently improves robustness\footnote{\ Some TQA work defines robustness as handling diverse temporal phenomena. Here, we define it as the ability to answer correctly despite suboptimal context}. We further validate our approach through a ``needle-in-the-haystack'' study, in which relevant context is buried among distractors. With forty distractors, RASTeR achieves 75\% accuracy, over 12\% ahead of the runner up.\footnote{\ Code available: https://github.com/danschumac1/RASTeR}

\end{abstract}

\section{Introduction}

Large language models (LLMs) can often answer factual questions directly from the knowledge stored in their parameters, if the necessary facts appeared in their pre-training data~\cite{petroni-etal-2019-language,roberts-etal-2020-much}. When a question is likely to require information outside of the model's pre-training data, practitioners typically fall back on retrieval‑augmented generation (RAG)~\cite{lewis2020retrieval}, which prepends retrieved passages to the prompt so the model can “read” before it “writes.” Unfortunately, the retriever offers no guarantee of relevance~\cite{yin-etal-2023-alcuna}. Irrelevant or adversarial snippets can mislead the generator and lower accuracy~\cite{petroni2020context}. Recent work further underscores that today’s QA benchmarks rarely stress a system’s robustness~\cite{shaier2024desiderata}. These issues become even more pronounced in TQA where answers depend on current facts and where stale, or simply wrong, documents are frequently retrieved~\cite{wu2024timesensitive}.

\begin{figure}[t]
    \centering
    \includegraphics[width=0.85\linewidth]{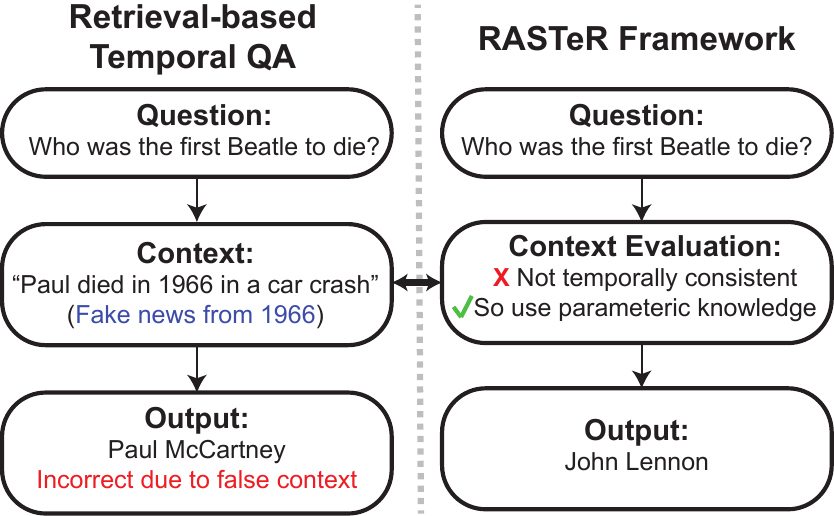}
    \caption{Example of TQA failure due to irrelevant context. The retrieved statement is outdated, leading to an incorrect answer. RASTeR detects the inconsistency and defaults to parametric knowledge. Additional experiments examine other context imperfections (e.g. partially incorrect, fully irrelevant context).}\vspace{-1em}
    \label{fig:demo}
\end{figure}

TQA, therefore, poses a dual challenge: models must identify any relevant entities and reason about their evolution over time. Robust temporal reasoning is critical for applications such as historical-event analysis~\cite{Lorenzini2022}, time-sensitive retrieval~\cite{wu2024timesensitive}, data-driven journalism, and real-time analytics, domains where a single date error can significantly alter the answer. Yet comprehensive benchmarks such as TimeBench and TRAM reveal that even GPT-4 lags behind human performance, despite access to gold context~\cite{chu2023timebench,wang2024tram}, and recent studies show that LLMs often hallucinate timelines or overlook explicit temporal cues~\cite{beniwal2024remember}. While prior work has evaluated models under clean context~\cite{wallat-etal-2025-study, luu-etal-2022-time, tan-etal-2024-towards}, tested zero-shot generalization with synthetic data~\cite{uddin-etal-2025-unseentimeqa}, or explored robustness to irrelevant context in general QA~\cite{yoran2024retrieval, cheng2024exploring}, these approaches do not directly address the temporal inconsistencies and ambiguities that arise in realistic retrieval.

Despite this progress, a key gap remains. Existing benchmarks and methods tend to focus on either (\textbf{1}) scenarios in which the model has no prior knowledge of the event and must rely entirely on external context, (\textbf{2}) general robustness to distractors without temporal grounding, or (\textbf{3}) evaluation of questions that may have been seen during pretraining. However, real-world TQA systems must handle both: reasoning about known events under noisy, outdated, or conflicting context, and generalizing to novel or emerging events where memorized knowledge offers no support. Temporal ambiguity~\cite{in-etal-2025-diversify, wang2025retrievalaugmentedgenerationconflictingevidence}, misaligned retrievals~\cite{fang2024enhancingnoiserobustnessretrievalaugmented, yoran2024retrieval}, and hallucinations even under gold context~\cite{he-etal-2024-retrieving, wallat2024temporal} highlight the need for methods that can diagnose and correct temporal inconsistencies. To our knowledge, no prior work systematically evaluates model robustness under both seen event settings and also evaluates unseen event settings with temporal context.

To address this gap, we propose RASTeR, an agenentic framework for TQA that explicitly separates context evaluation from answer generation. RASTeR introduces modular agents that assess the temporal relevance and coherence of retrieved passages before transforming valid evidence into a structured  TKG. This structure enables precise stepwise reasoning about time even under adversarial or outdated contexts. We systematically evaluate RASTeR across multiple models and four TQA datasets, including scenarios where events are known, unknown, or contextually distorted. Our results show that this agentic and structured decomposition not only improves robustness to noisy context, a key limitation in RAG pipelines, but also enables fine-grained reasoning over long, distractor-heavy passages. In doing so, RASTeR bridges the gap between robustness and temporal reasoning, offering a principled approach to TQA under realistic retrieval conditions. See Figure~\ref{fig:demo} for a high-level idea of our contribution.

Our contributions are as follows: (\textbf{1}) We introduce RASTeR, an agentic prompting pipeline that separates context evaluation from answer generation via temporal consistency agents and structured knowledge graph transformation. (\textbf{2}) We benchmark RASTeR across three LLMs and four TQA datasets, demonstrating consistent gains in both clean and noisy contexts. (\textbf{3}) We conduct granular robustness analyses, including adversarial retrieval settings (needle-in-the-haystack), altered temporal context, and relevance misclassifications, to better understand the strengths and weaknesses of this approach.

\section{Related Work}

\noindent \textbf{Temporal Question Answering.}
TQA tasks involve understanding how events unfold over time, whether in text, video~\cite{10.1007/s11263-017-1033-7}, or structured data such as knowledge bases~\cite{xiao2021next,jang2017tgif,saxena2021question,zhao2024utsa, tan-etal-2024-towards}. This includes applications like ordering clinical events~\cite{sun2013temporal, zhao2024utsa} or answering factoid questions such as ``Who was president of the U.S. in 1998?'' Several benchmarks have been proposed to evaluate temporal reasoning, including tests for time-sensitive fact verification and temporal reversal, where performance asymmetries between forward and backward questions reveal a reliance on memorized patterns rather than grounded temporal inference~\cite{bajpai-etal-2024-temporally, wallat-etal-2025-study}.

Recent and historical work has exposed persistent limitations of LLMs in this setting. Models often struggle to reason over timelines, hallucinate events, or miss temporal cues entirely~\cite{qiu2023large, beniwal2024remember}. To probe these weaknesses, researchers have released new datasets~\cite{gruber2024complextempqalargescaledatasetcomplex, 10.1145/3269206.3269247, velupillai-etal-2015-blulab, wang2022archivalqa, gruber2024complextempqa} and diagnostic tasks~\cite{llorens2015semeval, tan2023towards, gao2024two} that evaluate logical reasoning in time-sensitive settings. \citet{chenghaozhu-etal-2025-llm} highlight a related issue of temporal drift: LLMs tend to anchor their factual knowledge around 2015, resulting in degraded performance for domains like news or policy where timelines evolve. This drift presents a key challenge for retrieval-augmented QA, where the context retrieved may be outdated, misleading, or temporally misaligned with the question.

\vspace{2mm} \noindent \textbf{Robustness in Retrieval-Augmented Generation.}
Improving the robustness of LLMs to imperfect context has been a focus of recent work on RAG. Broadly, these methods fall into three categories: filtering irrelevant context, adversarial training, and ambiguity-aware reasoning. For filtering, \citet{yoran2024makingretrievalaugmentedlanguagemodels} propose using NLI-based filters to exclude unsupported evidence before generation, and fine-tune models on mixed-quality data to improve resistance to distractors. \citet{he-etal-2024-retrieving} introduce CoV-RAG, which incorporates a verification model and structured reasoning to select and integrate relevant information. More recently, \citet{chang2025mainrag} present MAIN-RAG, a multi-agent RAG framework where LLM agents collaboratively filter and score retrieved documents using adaptive, consensus-based thresholds to minimize noise without sacrificing recall. Similarly, \citet{nguyen2025marag} propose MA-RAG, which decomposes retrieval-augmented reasoning into specialized agent roles—Planner, Step Definer, Extractor, and QA—communicating through chain-of-thought prompting to iteratively refine retrieval and synthesis. In the legal domain, \citet{wang2025lmars} introduce L-MARS, a multi-agent workflow that coordinates reasoning, retrieval, and verification to reduce hallucination and uncertainty by decomposing legal queries, conducting targeted searches, and verifying jurisdictional validity before synthesis. While MAIN-RAG, MA-RAG, and L-MARS are all multi-agent systems that operate \textit{upstream} in the RAG pipeline to improve retrieval and evidence aggregation, \textit{our approach begins downstream}, assuming retrieval has already occurred and that the context mix is imperfect. In practice, our method is complementary, as both downstream and upstream methods could work in tandem  addressing a different phase of the reasoning process.

Adversarial training methods expose models to noisy or counterfactual inputs to encourage robustness. For instance, \citet{fang2024enhancingnoiserobustnessretrievalaugmented} train models on irrelevant and contradictory passages to improve reliability under real-world retrieval errors. However, these approaches typically focus on general QA and do not account for temporal-specific failure modes. Ambiguity-aware pipelines offer a complementary strategy. \citet{in-etal-2025-diversify} retrieve diverse evidence to accommodate questions with multiple valid answers. \citet{wang2025retrievalaugmentedgenerationconflictingevidence} propose a multi-agent architecture where separate models handle different retrieved passages, and a judge model resolves conflicts. Other work uses search-based methods~\cite{hu2025mctsragenhancingretrievalaugmentedgeneration}, eligibility assessment~\cite{kim-etal-2024-aligning}, or similarity-based example selection~\cite{park2024enhancingrobustnessretrievalaugmentedlanguage} to guide reasoning under ambiguity. Finally, GraphRAG~\cite{han2025retrievalaugmentedgenerationgraphsgraphrag} combines RAG with graph-structured knowledge, showing that graph-based retrieval can improve reasoning. This motivates us to explore how transforming retrieved temporal context into graph form to can support more robust reasoning.

\vspace{2mm} \noindent \textbf{Structured Knowledge and Reasoning.}
Structured representations such  TKGs enable reasoning over time.  Most prior research assumes access to structured datasets and focuses on TKG question answering (TKGQA), which typically involves either interpolation (inferring missing facts within a timeline) or extrapolation (predicting events beyond observed data)~\cite{chen-etal-2024-unified}. A central challenge in TKGQA is identifying the most salient nodes. \citet{zhang2024knowgptknowledgegraphbased} use reinforcement learning to sample reasoning chains, while \citet{gao2024twostagegenerativequestionanswering} first filter relevant relations and then restrict them temporally.

Others focus on improving question formulation. \citet{qianyihu-etal-2025-time} show that LLMs perform better on explicit temporal queries and propose a two-stage retrieval-and-rewriting pipeline to make implicit questions more solvable. \citet{xia-etal-2022-metatkg} also advocate for a two-step strategy that first retrieves direct evidence and then expands it using related entities to capture second-order temporal relationships. These methods assume relatively clean data and often ignore the noisy, conflicting nature of real-world context.

In contrast, our work examines how structured temporal representations impact model robustness when the context is messy, misaligned, or adversarial. Rather than using TKGs solely for interpolation or extrapolation, we dynamically construct TKGs from retrieved text and assess their utility under imperfect retrieval conditions. The closest to our approach is the Chain-of-Timeline framework~\cite{wu2025chainoftimeline}, which constructs structured TKGs based on a question and its associated context. However, their work evaluates only on golden context and a single dataset. We extend it by developing a system that handles a variety of context and generalizes across models and datasets.

\section{Method}

\begin{figure*}[t]
    \centering
      \includegraphics[width=\linewidth]{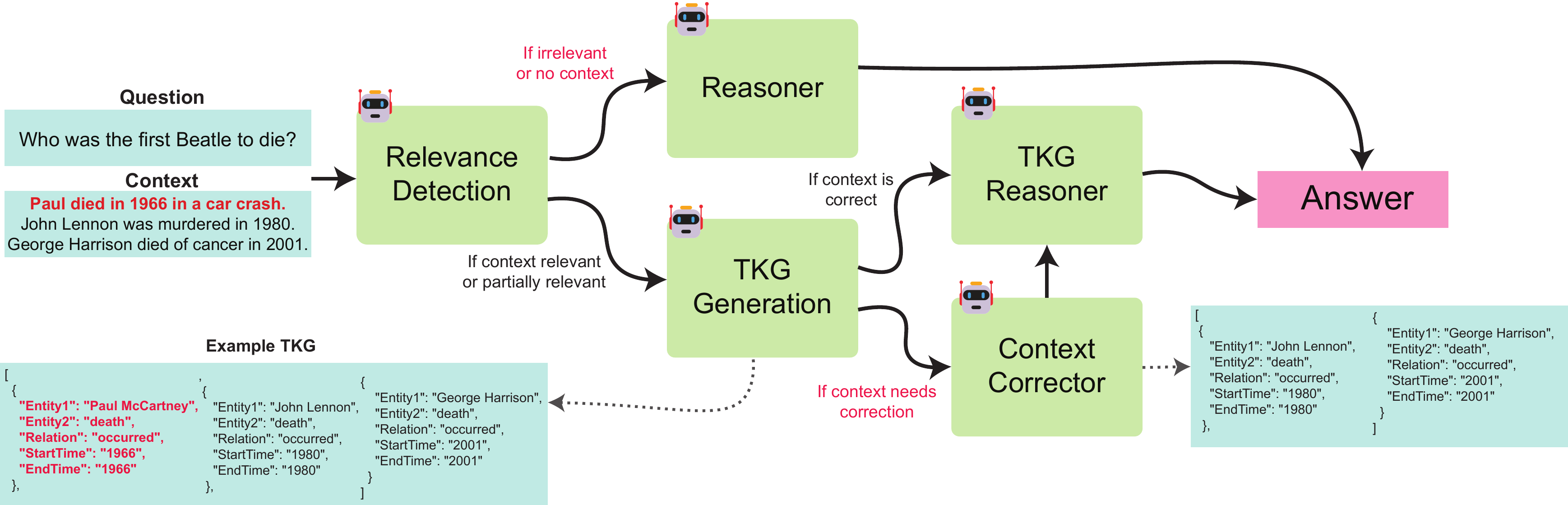}
    
    \caption{Overview of the RASTeR framework. Given a question and retrieved context, the system first determines whether the context is relevant and temporally coherent. If necessary, it corrects temporal inconsistencies before generating a structured TKG. The final answer is produced either by reasoning over the TKG or, in cases of irrelevant or missing context, via a fallback zero-shot reasoner.} \vspace{-1em}
    \label{fig:overview}
\end{figure*}

TQA presents unique challenges that standard RAG pipelines are not designed to handle. Retrieved context may be outdated, partially relevant, or temporally inconsistent, yet current systems often assume that any retrieved passage can be treated as reliable input. Our approach addresses this gap by explicitly separating context evaluation from answer generation. We first assess whether the context is relevant and temporally coherent with respect to the question. When the context passes these checks, we convert it into a structured TKG to support precise, time-aware reasoning. If the context is found to be unreliable or inconsistent, we either attempt to correct it or disregard it and rely on the model's parametric knowledge. This modular, agent-based design enables robust performance across a wide range of TQA scenarios through structured intermediate representations.

We formalize the robust TQA task as follows. Let $Q$ denote a temporal question, and $C$ denote the corresponding golden context to answer it. The LLM is modeled as a function $\mathcal{M}_\theta$ that produces an answer $A = \mathcal{M}_\theta(Q, C)$, where $Q = {q_1, \dots, q_n}$, $C = {c_1, \dots, c_m}$, and $A = {a_1, \dots, a_k}$ are token sequences. We evaluate model performance across several context settings: relevant ($C_r$), irrelevant ($C_i$), altered ($C_a$), and no context ($C_0$).

The RASTeR pipeline is structured into distinct modules, each handled by a dedicated agent: context evaluation, TKG construction, context correction, and answer generation (via reasoner agents).\footnote{\ Multi-agent frameworks generally increase token usage. See Appendix~\ref{appendix:Estimated_Token_Costs} for a detailed breakdown of RASTeR’s computational cost.} We show a high-level overview of our method in Figure~\ref{fig:overview}. First, a question and context are evaluated to test if the context is relevant to the question. If not, or if there is no context, a reasoner answers the question directly using the models internal parametric without using the context. Otherwise, a TKG is generated using the context. If the relevance detector determined that the context was only partially relevant, then the context corrector is called to fix the TKG. Finally, the TKG reasoner is called to reason over the TKG to answer the question.  We describe each part below.

\vspace{2mm} \noindent \textbf{Context Evaluation.}
Before reasoning over the retrieved context, we must establish whether it is temporally aligned and semantically relevant to the question. To achieve this, we introduce a Relevance Reasoning Chain that decomposes context evaluation into discrete steps. Given a question $Q$ and context $C$, the model identifies the question's entities $Q_e$, checks for their presence ($e_{\text{pres}}$) in the context, and generates a Correction Reasoning Chain $D = (d_1, d_2, d_3, d_4)$ assessing $d_1$: chronological coherence of dates, $d_2$: alignment of context dates with the question, $d_3$: realism of the overall time span,  and $d_4$: agreement with parametric knowledge.
These outputs inform a final decision $C_{\text{nc}}$ on whether the context requires correction. If $C_{\text{nc}} = True$, then the Context Correction agent (details below) is triggered to modify the context. The exact prompting format as well as an example for this step is shown in Appendix~\ref{appendix:prompts:RASTeR} Figure~\ref{fig:relevance_prompt} and Figure~\ref{fig:relevance_agent_example} respectively.

\vspace{2mm} \noindent \textbf{Temporal Knowledge Graph Construction.}
When the context is deemed usable, we convert it into a TKG that supports symbolic reasoning over events and temporal intervals. TKGs can be formally defined as a sequence of quadruples ${(e_1, r_i, e_2, t_i)}_{i=1}^N$, where each tuple represents a fact consisting of a subject entity $e_i$, a relation $r_i$, an object entity $e_2$, and an associated timestamp $t_i$.. We begin by splitting the context into sentences and chunking it with overlap (batch size = 12, overlap = 6). For each chunk $c_i$, the model conditions on the previous TKG state $TKG_{i-1}$ to expand the graph: $TKG_i = f_{\text{TKG}}(c_i, TKG_{i-1})$. The final graph is the union of all iterations. As an example, if there are three sentences, $s_1, s_2$, and $s_3$, with a batch size of 2 and an overlap of 1,  a $TKG_1$ will be generated using $s_1$ and $s_2$. $TKG_2$ will be generated by $s_2$ and $s_3$. Both $TKG_1$ and $TKG_2$ will be combined to form the final $TKG$. Intuitively, generating a TKG by passing the entire context as input causes the model to hallucinate nodes and edges, and worse, miss important information. By generating it in an iterative and overlapping fashion, information is seen multiple times and in small contexts to generate the final graph better.  An example of the prompting for this procedure is shown in Appendix~\ref{appendix:prompts:RASTeR} Figure~\ref{fig:tkg_constructor_prompt}.

\vspace{2mm} \noindent \textbf{Context Correction.}
If $C_{\text{nc}}$ is true, we trigger a context correction mechanism. For each TKG node, we replace the temporal fields (\texttt{starttime} and \texttt{endtime}) with placeholders and prompt the model to infer plausible time spans. The model then generates a natural language sentence that articulates the relation. Formally, each corrected node is $(e_1, \textit{rel}, e_2, \textit{starttime}_i', \textit{endtime}_i', \textit{sentence}_i)$. This results in a corrected graph $TKG'$ with both symbolic and textual representations. Appendix~\ref{appendix:prompts:RASTeR} Figure~\ref{fig:tkg_corrector_prompt} shows the full correction prompt.

\vspace{2mm} \noindent \textbf{Answer Generation.}
When a TKG is available, we filter relevant nodes, extract justifications, and synthesize an answer $A$ as part of a larger output $(A, sn, r) = f_{\text{tkg\_answerer}}(Q, TKG)$, where $sn$ denotes supporting nodes and $r$ is the reasoning trace. We do this using an LLM agent without any rule-based methods. The full answer generation prompt is shown in Appendix~\ref{appendix:prompts:RASTeR} Figure~\ref{fig:tkg_reasoning}.
In the absence of usable context (without TKG), the model falls back to zero-shot reasoning using parametric knowledge: $(Q^1, r, A) = f_{\text{zs\_answerer}}(Q)$, where $Q^1$ is a restated version of the question and $r$ is the internal reasoning trace. The prompt for this setup is included in Appendix~\ref{appendix:prompts:RASTeR} Figure~\ref{fig:no_tkg_reasoning}.

\section{Experiments}

In this section, we describe the datasets, metrics, baseliens, and our overall results. We also include a detailed error analysis and ablation of the various components in our agent-based framework.

\vspace{2mm} \noindent \textbf{Datasets.}
Each subset of our collected datasets benchmarks a distinct aspect of temporal reasoning, thus testing different dimensions of temporal question answering. We describe each dataset below.

\vspace{2mm} \noindent \textbf{\textit{MenatQA (MQA).}}
In MQA~\cite{wei-etal-2023-menatqa}, the \emph{counterfactual} subset explores imaginative temporal reasoning. The \emph{scope} subset evaluates a model’s ability to handle questions with variable time spans, while the \emph{scope\_expand} subset challenges models to reason over extended temporal intervals that go beyond the typical bounds of the context. The \emph{order} subset targets reasoning over shuffled event sequences.

\vspace{2mm} \noindent \textbf{\textit{TimeSensitiveQA (TSQA).}}
TSQA~\cite{TimeSensitiveQA} evaluates temporal reasoning over time-evolving passages, with a focus on alignment between temporal expressions in the question and timeline boundaries in the context. The dataset is split into two levels: \emph{easy} and \emph{hard}. In the \emph{easy} subset, the time specifier in the question exactly matches a boundary event (e.g., the start or end of a time interval) that is explicitly mentioned in the passage, allowing models to answer via surface-level matching. In the \emph{hard} subset, the time specifier falls within the middle of a temporal span, requiring models to infer implicit time alignment and reason beyond explicit timestamps.

\vspace{2mm} \noindent \textbf{\textit{TempReason (TR).}}
TR~\cite{tan-etal-2023-towards} focuses on factual temporal reasoning across two levels. The \emph{l2} subset asks for specific facts grounded in time (e.g., “Who coached the team in 2010?”), while the \emph{l3} subset requires reasoning over event sequences (e.g., “Who coached the team before Ted Lasso?”), combining time understanding with knowledge of event order.

\vspace{2mm} \noindent \textbf{\textit{UnSeenTimeQA (UTQA).}}
UTQA~\cite{uddin-etal-2025-unseentimeqa} is a dataset of logistics-style word problems designed to test temporal reasoning without relying on prior knowledge. Because the problems are synthetic and domain-specific, models cannot answer them without using the provided context. This reduces concerns about training data contamination. We focus on the \emph{hardSerial} and \emph{hardParallel} subsets. \emph{HardSerial} assumes events occur in sequence but only provides durations, requiring models to simulate a timeline mentally. \emph{HardParallel} allows events to overlap and introduces distractors that resemble irrelevant but plausible context.

\vspace{2mm} \noindent \textbf{Data Augmentation.}
Out of the box, each dataset consists of \( N \) rows containing \( Q \), \( A \), and \( C_r \). Formally, \(\mathcal{D} = \{(Q_n, A_n, C_r^{(n)})\}_{n=1}^{N}\). Our goal is to augment each row of each dataset\footnote{\ We do not augment UTQA because its questions cannot be answered without the provided golden context (e.g., “Package-A arrived from Location-1 to Location-2 on Plane B at 4:00 pm”). In contrast, the other datasets contain questions (e.g., “Where did Messi play before Miami?”) that large language models can often answer correctly from pretrained knowledge, even without golden context.}
 \( n \in \{1, \dots, N\} \) of \(\mathcal{D}\) with \( C_i \) and \( C_a \), yielding the extended dataset \(\mathcal{D}' = \{(Q_n, A_n, C_r^{(n)}, C_i^{(n)}, C_a^{(n)})\}_{n=1}^{N}\).

\(C_a\) is constructed from \(C_r\) by (1) using regex to identify all explicit temporal expressions and (2) applying a rule-based substitution that replaces each temporal expression (e.g., “January,” “Jan,” “1994,” “01-1995”) with a different, non-matching value; i.e., \(C_a^{(n)} = \tau\!\big(C_r^{(n)}\big)\), where \(\tau\) maps each temporal token \(t\) to \(\tilde{t} \neq t\).

To generate \(C_i\) within the same dataset, we randomly sample another row’s \(C_r\): $
\mathcal{C}_i^{(n)} = C_r^{(m)}, \quad \text{where  }\,\, m \sim \mathrm{Uniform}(\{1, \dots, N\} \mid C_r^{(m)} \neq C_r^{(n)}).
$

In realistic RAG scenarios, retrieved documents are rarely \textit{completely} irrelevant. However, each dataset already contains highly related questions (for example, TR primarily consists of \(Q\) and \(C\) pairs about athletes’ careers), so this setup is sufficient to approximate realistic retrieval noise. To further validate the robustness of our system, we also evaluate RASTeR under conditions where the retrieved context is not random but the \textit{most semantically similar} to the query. See Appendix~\ref{appendix:semantically_similar_context}.

\vspace{2mm} \noindent \textbf{Metrics.}
We report Exact Match (EM), Contains Accuracy (Acc), and word-level F1 to evaluate model performance. EM measures whether the predicted answer exactly matches any reference answer (e.g ``Barack Obama'' is not ``Obama''). Acc is more lenient and considers a prediction correct if it is a subset of, or contains, any reference answer (e.g ``Barack Obama'' contains ``Obama''). Finally, F1 captures the overlap between the predicted and reference answers at the word level by computing the harmonic mean of precision and recall. Formal definitions of these metrics are provided in Appendix~\ref{appendix:metrics}. To conserve space, our main tables only show Acc. The full results which include EM and F1 are available in Appendix~\ref{appendix:Expanded Results} (Tables~\ref{tab:em_avg} and~\ref{tab:f1_avg}).

\vspace{2mm} \noindent \textbf{Baselines.}
We compare three baseline prompting strategies against our proposed method. (\textbf{1}) generic Few-Shot prompting, (\textbf{2}) a simple reasoning prompt, and (\textbf{3}) a TKG prompt with no agentic steps. For each baseline we include four few-shot examples, one each for relevant-, irrelevant-, slightly altered, and no-context.

\vspace{2mm} \noindent \textbf{\textit{Few-Shot}.}
In the Few-Shot approach, we provide the question and context and ask for an answer. The prompt for this baseline is in Figure~\ref{fig:few-shot_prompt} in the Appendix.

\vspace{2mm} \noindent \textbf{\textit{Reasoning}.}
In the reasoning approach, we ask the model to follow the following reasoning chain (\textbf{1}) restate the question, (\textbf{2}) evaluate the relevance of the context, (\textbf{3}) quote supporting evidence, (\textbf{4}) reason towards an answer, and (e) use the reasoning to come to a final conclusion. Basically, this is a structured chain-of-thought-like prompt~\cite{sultan2024structured} for TQA. The full prompt can be seen in Figure~\ref{fig:Reasoning} in the Appendix.

\vspace{2mm} \noindent \textbf{\textit{Simple TKG}.}
In this approach, the model first extracts entities from the context and uses them to construct a structured TKG composed of time-stamped relational tuples. It then answers the question using only the generated TKG, encouraging structured reasoning and temporal grounding without additional agentic steps.
Unlike the simple TKG baseline, which directly constructs a TKG from the context without evaluating its relevance or consistency, our method introduces agentic reasoning steps. These include checking whether the context is relevant or altered, correcting temporal inconsistencies, and iteratively building a TKG conditioned on previous outputs, resulting in a more robust and context-sensitive reasoning process. The full prompt is in Figure~\ref{fig:simple_tkg} in the Appendix.

%%%%%%%%%%%%%%%%%%%%%%%%%%%%%%%%%%%%%%%%%%%%%%%%%%%%%%%%%%%%%%%%%%%%%%%%%%%%%%%%%%%%%%%%%%%%%%%%%%%%%%
%%%%%%%%%%%%%%%%%%%%%%%%%%%%%%%%%%%%%%%%%%%%%%%%%%%%%%%%%%%%%%%%%%%%%%%%%%%%%%%%%%%%%%%%%%%%%%%%%%%%%%

\begin{table}[t]
\centering
\resizebox{\linewidth}{!}{
\begin{tabular}{llcccc}
\toprule
\textbf{Model} & \textbf{Prompt Type} & \textbf{MQA} & \textbf{TR} & \textbf{TSQA} & \textbf{Avg} \\
\midrule
\multirow{4}{*}{\textbf{gemma-3-12b-it}}& Few-Shot&     \textbf{.332}& .257& .176& .293
\\
& Reasoning&    .222& .275& \textbf{.190}& .225
\\ \cmidrule(lr){2-6}
& TKG&          .302& .254& .164& .271
\\
& RASTeR&     .327& \textbf{.290}& .166& \textbf{.294}
\\
\midrule
\multirow{4}{*}{\textbf{gpt-4o-mini}}& Few-Shot&     .302& .288& .220& .286
\\
& Reasoning&    .264& \textbf{.324}& .236& .270
\\ \cmidrule(lr){2-6}
& TKG&          .306& .256& .201& .280
\\
& RASTeR&     \textbf{.319}& .315& \textbf{.262}& \textbf{.311}
\\
\midrule
\multirow{4}{*}{\textbf{Llama-3.1-8B-Instruct}} 
& Few-Shot&     .087& .124& .069& .090
\\
& Reasoning&    .217& .227& .135& .205
\\ \cmidrule(lr){2-6}
& TKG&         \textbf{ .266}& .227& .135& \textbf{.238}\\
& RASTeR&     .253& \textbf{.231}& \textbf{.182}& \textbf{.238}\\
\bottomrule
\end{tabular}}

\caption{Acc averaged across subset, and eval-context for each model and prompting strategy.} \vspace{-1em}
\label{tab:cont_acc_avg}
\end{table}

\section{Results}
\vspace{2mm} \noindent \textbf{Main Results.} Table~\ref{tab:cont_acc_avg} shows the average contains accuracy on the MQA, TR, and TSQA datasets. RASTeR demonstrates consistent robustness across MQA, TR, and TSQA. It generalizes well across Gemma (gemma-3-12b-it) , GPT (gpt-4o-mini) , and Llama (Llama-3.1-8B-Instruct) with an average improvement in accuracy from .293, .286, and .205 to .294, .311, and .238, respectively. These scores are averaged across all four context types: relevant ($C_r$), irrelevant $C_i$, altered ($C_a$), and no context ($C_0$). For Gemma and LLaMA, TKG ties for best average score. Overall, this shows strong robustness to noisy RAG contexts compared to standard baseline methods. Significance testing further validates these results; details can be found in Appendix \ref{appendix:significance_testing}

%%%%%%%%%%%%%%%%%%%%%%%%%%%%%%%%%%%%%%%%%%%%%%%%%%%%%%%%%%%%%%%%%%%%%%%%%%%%%%%%%%%%%%%%%%%%%%%%%%%%%%

\begin{table}[t]
\centering
\resizebox{\linewidth}{!}{
\begin{tabular}{llcccc}
\toprule
\textbf{Model} & \textbf{Prompt} & \textbf{MQA} & \textbf{TR} & \textbf{TSQA} & \textbf{Avg} \\
\midrule
\multirow{4}{*}{\textbf{gemma-3-12b-it}} 
& Few-Shot
& .238& .004& .010& .161
\\
& Reasoning& .110& .026& .028& .082
\\ \cmidrule(lr){2-6}
& TKG
& .235& .004& .010& .159
\\
& RASTeR      
& \textbf{.305}& \textbf{.052}& \textbf{.098}& \textbf{.228}
\\
\midrule
\multirow{4}{*}{\textbf{gpt-4o-mini}} 
& Few-Shot
& .190& .018& .044& .137
\\
& Reasoning& .174& \textbf{.116}& .120& .155
\\ \cmidrule(lr){2-6}
& TKG
& .211& .014& .030& .148
\\
& RASTeR      
& \textbf{.253}& .091& \textbf{.164}& \textbf{.171}
\\
\midrule
\multirow{4}{*}{\textbf{Llama-3.1-8B-Instruct}} 
& Few-Shot
&                   .019& .000& .002& .013
\\
& Reasoning&                   .090& .000& .002& .060
\\ \cmidrule(lr){2-6}
& TKG
&                   .179& .012& .018& .124
\\
& RASTeR&         \textbf{.209}& \textbf{.050}& \textbf{.102}& \textbf{.165}\\
\bottomrule
\end{tabular}}
\caption{Acc averaged across subset for Irrelevant Context Evaluations Only.} \vspace{-1em}
\label{tab:cont_acc_random}

\end{table}

Next, in Table~\ref{tab:cont_acc_random}, we report how our system works in a worst-case setting: when evaluated only on  irrelevant context.  On average, RASTeR consistently outperforms other methods, particularly on open-source models. RASTeR with Gemma scores on average $\sim$7\% better (.228) than the runner-up (.161). Likewise, LLaMA (.165) handles random context on average $\sim$6\%  better than its runner-up (.124). Furthermore, in the irrelevant evaluation setting, our method is the \textbf{dominant prompting strategy across nearly every dataset and model combo}. The only exceptions being gpt + TR, where reasoning is higher RASTeR (.116 vs. .091) 

%%%%%%%%%%%%%%%%%%%%%%%%%%%%%%%%%%%%%%%%%%%%%%%%%%%%%%%%%%%%%%%%%%%%%%%%%%%%%%%%%%%%%%%%%%%%%%%%%%%%%%
\vspace{2mm} \noindent \textbf{Needle-In-The-Haystack.} In practice, RAG systems often surface both relevant and irrelevant content. The context is \textit{generally} never completely relevant nor completely irrelevant. To simulate this, we manipulate TSQA by inserting $n$ irrelevant contexts on each side of the golden context (e.g., for $n=3$: \textit{irr}, \textit{irr}, \textit{irr}, \textit{rel}, \textit{irr}, \textit{irr}, \textit{irr}; where `\textit{irr}' is an unrelated distractor and `\textit{rel}' is the true relevant context). Intuitively, the model needs to identify the relevant context within many noisy contexts. This is particularly difficult given that language models generally ``lose'' information in the middle~\cite{liu2024lost,zhang2025lost}. Descriptive statistics for this experiment can be found in  Figure~\ref{tab:discriptive_statistics} in the Appendix.  Figure~\ref{fig:haystack} shows the overall findings of our experiments. Overall, RASTeR remains highly effective under this setup, maintaining strong performance despite the presence of distractors. At each $n$, RASTeR achieves the highest performance. In fact, with forty distractors ($n=20$), RASTeR, with an accuracy of 74\%, outperforms (by at least 12\%) all other prompting strategies' performance at fourteen distractors ($n=7$). These results demonstrate that our system can robustly reason over long contexts with numerous distractors. This finding is highly impactful, given that we show that careful engineering of how context is handled, even at a low, nearly artificial level, can generalize to more realistic scenarios that are experienced in practice.

\begin{figure}[t]
    \centering
    \includegraphics[width=.9\linewidth]{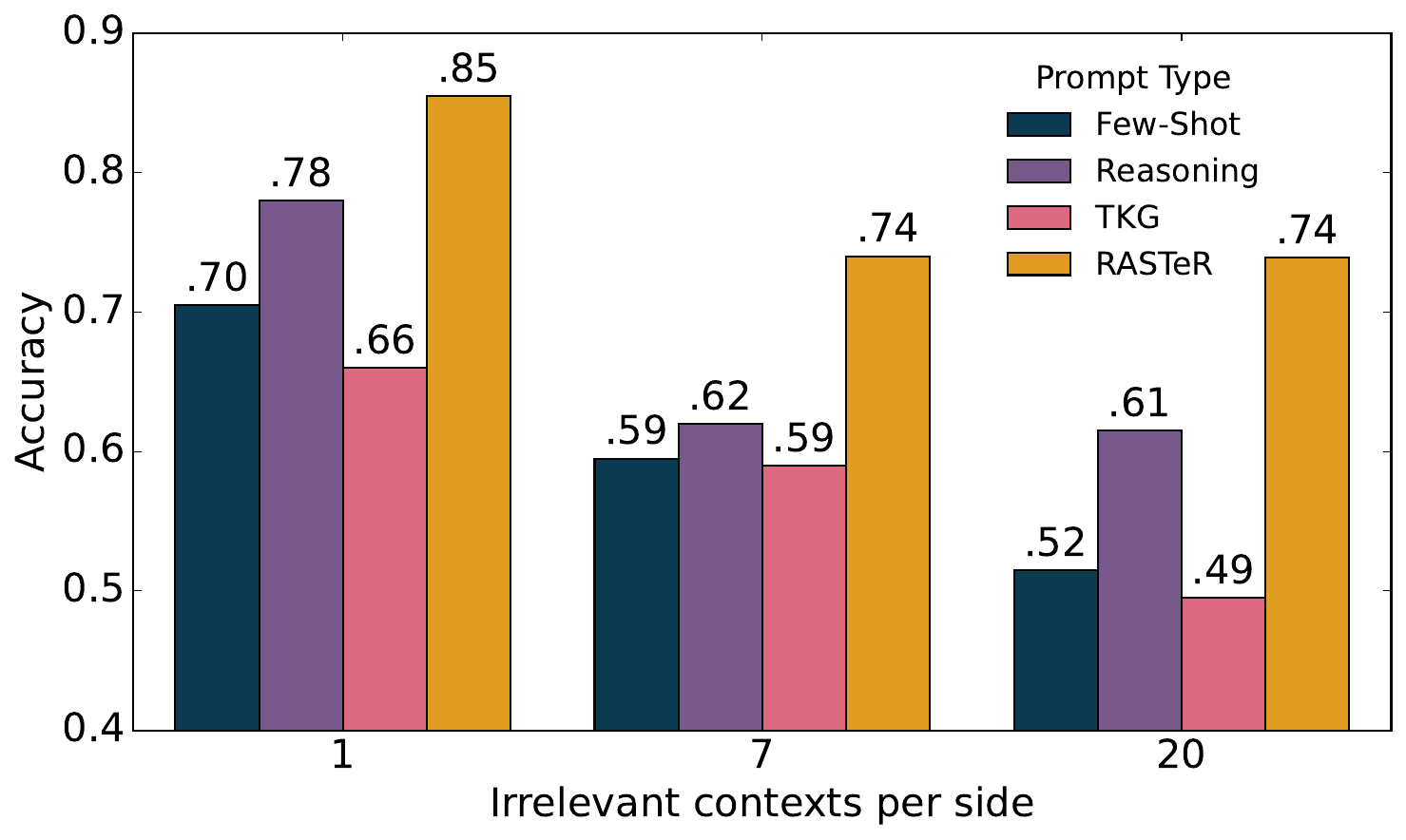}
\caption{GPT accuracy as the number of distractors (irrelevant contexts)  increases around a single relevant passage. All contexts have a relevant passage.} \vspace{-1em}
    \label{fig:haystack}
\end{figure}

%%%%%%%%%%%%%%%%%%%%%%%%%%%%%%%%%%%%%%%%%%%%%%%%%%%%%%%%%%%%%%%%%%%%%%%%%%%%%%%%%%%%%%%%%%%%%%%%%%%%%%

\vspace{2mm} \noindent \textbf{Unseen Data.} In Table~\ref{tab:unseen}, we report the results of the USQA dataset. Intuitively, we are evaluating generalization to unseen data, i.e., information the model has never seen during pretraining. At a high level, we hypothesize that using the TKG is crucial for improved temporal reasoning when the temporal context wasn't observed during pretraining. While RASTeR incorporates a TKG, it may not consistently outperform the TKG baseline alone, as RASTeR's reasoning and TKG components are decoupled to better handle noisy context. In contrast, the TKG baseline reasons and answers within a single prompt. We find that RASTeR outperforms the reasoning and few-shot baselines across all models and metrics, confirming that incorporating a TKG, even in a modular setup, substantially enhances generalization to novel temporal contexts. For instance, using the gemma-3-12b-it model, RASTeR achieves an average accuracy of 0.373 compared to only 0.149 for the reasoning baseline and 0.101 for few-shot prompting. This trend holds across other models as well, such as LLaMA, where RASTeR improves from 0.120 (few-shot) and 0.195 (reasoning) to 0.235. Although RASTeR does not always exceed the decoupled TKG baseline, its consistent advantage over non-TKG methods demonstrates the importance of explicitly structured temporal representations even in modular reasoning pipelines. This result suggests that future work can explore how to better link the reasoning answerer and the actual TKG generation (e.g., through iterative TKG generation and answering, in a back-and-forth framework).

\begin{table}[t]
\centering
\resizebox{\linewidth}{!}{
\begin{tabular}{llccc}
\toprule
\textbf{Model} & \textbf{Prompt Type} & \textbf{HardParallel} & \textbf{HardSerial} & \textbf{Avg} \\
\midrule
\multirow{4}{*}{\textbf{gemma-3-12b-it}}& Few-Shot&         .085& .117& .101
\\
& Reasoning&        .146& .151& .149
\\  \cmidrule(lr){2-5}
& TKG&              \textbf{.341} & \textbf{.408}& \textbf{.375}
\\
& RASTeR      &   \textbf{.391}& \textbf{.355}& \textbf{.373}
\\
\midrule
\multirow{3}{*}{\textbf{gpt-4o-mini}}& Few-Shot
&                   .267& .317& .292
\\
& Reasoning
&                   .190& .164& .177
\\  \cmidrule(lr){2-5}
& TKG
&                   \textbf{.533}&\textbf{ .551}& \textbf{.542}
\\
 & RASTeR      &  \textbf{.251} & \textbf{.331} & \textbf{.291}
\\
\midrule
\multirow{4}{*}{\textbf{Llama-3.1-8B-Instruct}} 
& Few-Shot
&                   .109& .130& .120
\\
& Reasoning
&                   .197& .192& .195
\\  \cmidrule(lr){2-5}
& TKG
&                   \textbf{.275}& \textbf{.274}& \textbf{.275}
\\
& RASTeR      &   \textbf{.213} & \textbf{.256}& \textbf{.235}\\
\bottomrule
\end{tabular}}
\caption{Acc across the UTQA hard subsets using relevant context only.}
\label{tab:unseen}
\end{table}

\begin{table}[t]
\centering
\resizebox{.6\linewidth}{!}{
\begin{tabular}{lcc}
\toprule
Ablation & \textbf{irrelevant} & \textbf{avg} \\
\midrule
RASTeR     & \textbf{.300} & \textbf{.388} \\ \midrule
w/o DateFix         & .263 & .375 \\
w/o TKG             & .275 & .360 \\
w/o DetRel          & .212 & .325 \\
\bottomrule
\end{tabular}}
\caption{Ablation results showing \textit{irrelevant} accuracy context and the overall average across all context types.} \vspace{-1em}
\label{tab:ablationmain}
\end{table}

\vspace{2mm} \noindent \textbf{Ablations.}
RASTeR, like all multi-step systems, introduces additional potential failure points at each agentic stage, which can increase the risk of cascading errors. While there are no hard-coded issues that cause complete breakdowns—each module can successfully pass its output to the next—individual components may still produce suboptimal results. For instance, the relevance agent might incorrectly flag a relevant context as needing correction, or the TKG generator could hallucinate or omit critical temporal facts. Despite these possibilities, our results show that the overall framework consistently improves performance across models and datasets. To better understand these dynamics, we perform ablation studies and manual analyses to evaluate the contribution of each component and identify areas where further improvement is possible.

To assess the contribution of each component in RASTeR, we conducted an ablation study by evaluating three modified variants of the pipeline: (\textbf{1}) \textit{w/o DateFix}, which disables the context corrector responsible for resolving temporally inconsistent information; (\textbf{2}) \textit{w/o TKG}, which removes the TKG constructor and relies entirely on natural language rather than structured graphs; and (\textbf{3}) \textit{w/o DetRel}, which bypasses relevance assessment by treating all input context as relevant. Each ablation was tested against the full pipeline on a randomly sampled subset spanning all datasets and subsets. Descriptive statistics for this subset appear in Table~\ref{tab:ablation_descriptive_statistics} (Appendix~\ref{appendix:Descriptive Statistics}]). Overall, the full RASTeR pipeline achieves the highest average accuracy (.388), outperforming all ablations. In the irrelevant context setting, RASTeR also obtains the best performance (.300), indicating that both the TKG and context relevance agents contribute meaningfully to robustness under noisy retrieval. Notably, removing the relevance detector (\textit{w/o DetRel}) leads to the largest drop in performance, especially in the irrelevant context setting, suggesting that misclassifying noisy inputs as relevant can significantly degrade reasoning. These results highlight the importance of both structured temporal representation and selective context filtering in improving TQA robustness. Full ablation results are shown in Table~\ref{tab:ablation} (Appendix~\ref{appendix:ablation}).
\begin{table}[t]
\centering
\resizebox{0.95\linewidth}{!}{
\begin{tabular}{llccc}
\toprule
\multicolumn{2}{c}{} & \multicolumn{3}{c}{\textbf{Predicted Context Type}} \\
\cmidrule(lr){3-5}
& & \textbf{SA} & \textbf{NO/IRR}& \textbf{REL}\\
\midrule
\multirow{4}{*}{\textbf{Eval Context}}& \texttt{No}                & 0.0\%   & 100.0\% & 0.0\%   \\
& \texttt{Irrelevant}        & 2.3\%   & 90.1\%  & 7.6\%   \\
& \texttt{Relevant}          & 2.6\%   & 4.8\%   & 92.6\%  \\
& \texttt{Slightly Altered}  & 77.1\%  & 8.5\%   & 14.3\%  \\
\bottomrule
\end{tabular}
}
\caption{Confusion matrix of RASTeR’s predicted context type versus true context type. Results are aggregated across TR, MQA, and TQA.}
\label{tab:relevance_pred}
\vspace{-1em}
\end{table}

\vspace{2mm} \noindent \textbf{Validating Relevance Determiner.} 
In Table~\ref{tab:relevance_pred}, we observe distinct patterns in how relevance is predicted across different eval contexts. When evaluated with no context, the model perfectly classifies the context as \texttt{NO / IRR}, 100\% of the time. When presented with irrelevant context, the model sometimes mistakenly labels  the context as \texttt{REL} (7.6\%) or less frequently as \texttt{SA} (2.3\%) . This is followed by evaluating on relevant context, in which the model most frequently misclassified the context as \texttt{NO / IRR} (4.8\% of the time). There is a big drop in performance when evaluated on \texttt{slightly altered}, relevance errors are more evenly split. While 72.6\% of predictions are correctly labeled as \texttt{SA}, the 14.3\% mislabeled as \texttt{REL} and 8.5\% as \textit{NO / IRR} suggest that identifying slightly altered context remains challenging and leaves room for improvement. UTQA is not include in the aggregation in Table \ref{tab:relevance_pred}. UTQA only contains relevant context and aggregating a cross it would skew results. With that being said, RASTeR identifies UTQA's relevant context as relevant 100\% of the time.

\vspace{2mm} \noindent \textbf{Validating TKG.}
Another potential source of error lies in the temporal knowledge graph (TKG) construction process. If the constructor fails to accurately identify entities, relations, or timestamps, the resulting graph would likely be less informative than the original relevant context. Although the overall performance gains (in EM, contains accuracy, and F1) suggest that this is not a major concern, we conducted an additional validation to confirm. Two annotators manually evaluated 365 extracted triples sampled from each dataset (\textit{entities, relations, timestamps}), labeling each as fully correct, partially correct, or incorrect. The annotators showed strong agreement, assigning the same label in 86\% of cases. Of the 365 triples, 297 (81.37\%) were fully correct, 51 (13.97\%) incorrect, and 17 (4.66\%) partially correct. Disagreements were jointly reviewed and reconciled by both annotators.

\vspace{2mm} \noindent \textbf{{Types of General Errors.}}
A common issue occurs when the model attempts to infer an answer even when the gold label is unanswerable. 
\begin{center}
    \centering
\footnotesize 
\begin{tcolorbox}[colback=gray!5, colframe=gray!75!black, width=0.95\linewidth, sharp corners=south, boxsep=1pt, left=2pt, right=2pt, top=2pt, bottom=2pt]
\textbf{QUESTION:} What job did Mary have in 2010? \\
\textbf{CONTEXT:} In 2009, Mary was a teacher at Lincoln High School. In 2011, she became a school principal. \\
\textbf{INCORRECT REASONING:} Mary's 2010 job is not explicitly stated. But since she was a teacher in 2009 and only became a principal in 2011, it is inferred she remained a teacher in 2010.
\textbf{GROUND TRUTH:} unanswerable
\textbf{PREDICTION:} Lincoln High School
\end{tcolorbox}
\end{center}

Our prompting pipeline encourages models to reason and guess in the absence of explicit evidence, which improves performance in no-context and irrelevant-context settings. However, this behavior can produce errors in settings where abstaining is preferred. An example is provided above.

Finally, temporal reasoning remains one of the most challenging categories of errors. To highlight these issues, MQA’s counterfactual questions require models to answer based on a hypothetical that directly contradicts the context. These questions test whether models adhere to the ``what-if'' condition rather than relying on factual timelines. example of a subset-specific reasoning error can be found below:
\begin{center}
    \centering
    \footnotesize 
\begin{tcolorbox}[colback=gray!5, colframe=gray!75!black, width=0.95\linewidth, sharp corners=south, boxsep=1pt, left=2pt, right=2pt, top=2pt, bottom=2pt]
\textbf{QUESTION:} What school did Henry go to from 2008 to 2010, if Henry didn’t graduate from Rice High School until 2011? \\
\textbf{CONTEXT:} Henry started at Rice High School in 2004. In 2008, he graduated and enrolled at Brown University. He completed his studies there in 2015. \\
\textbf{INCORRECT REASONING:} The timeline shows Henry enrolled at Brown University in 2008, which implies he attended it from 2008 to 2010. Since no other school is mentioned, Brown is inferred as the correct answer. \\
\textbf{GROUND TRUTH:} Rice High School \\
\textbf{PREDICTION:} Brown University
\end{tcolorbox}
\end{center}
These examples illustrate the need for finer-grained evaluation and improved handling of temporal and counterfactual reasoning in large language models.

\begin{comment}
    The main takeaways are as follows:  
(1) The full RASTeR pipeline achieves the highest average accuracy (.388), demonstrating the benefit of combining all agents.  
(2) Its performance is most dominant in the slightly altered (.250) and irrelevant (.300) context settings, where temporal correction and context assessment are most critical. Additionally, RASTeR nearly ties in the no-context setting, trailing the best ablation by just 1.2\%.
(3) In the relevant context condition, both the \textit{w/o DateFix} and \textit{w/o DetRel} ablations (which are functionally identical when evaluated on relevant context see Figure~\ref{fig:pipeline}) achieve the highest accuracy (.762), slightly outperforming the full pipeline. This is expected: as shown in Table~\ref{tab:relevance_pred}, RASTeR occasionally misclassifies relevant context as needing correction, introducing unnecessary changes and downstream errors.
\end{comment}

\section{Conclusion}
TQA presents persistent challenges for LLMs, particularly when retrieved context is irrelevant, misleading, or missing. We introduced RASTeR, a modular, agentic framework that separates context evaluation from answer generation. By assessing context quality, constructing structured TKGs, and correcting inconsistencies, RASTeR enables more robust and temporally grounded reasoning. Across four TQA datasets and three LLMs, RASTeR consistently improves accuracy in noisy and distractor-heavy settings while maintaining strong performance in ideal conditions. In needle-in-the-haystack scenarios, it not only outperforms alternatives but also degrades more gracefully as distractors increase.
In future work, we plan to extend RASTeR to support multi-hop temporal reasoning and questions with multiple temporally valid answers. We also aim to broaden our robustness analysis beyond date shifts to include perturbations such as entity substitutions and relation modifications, better characterizing model sensitivity to noisy temporal input. Furthermore, we aim to investigate how to more effectively integrate TKG generation and the reasoner answerer for improved performance on unseen temporal reasoning questions.

\section*{Acknowledgements}
This material is based upon work supported by
the National Science Foundation (NSF) under
Grant No. 2145357. This research was also partially sponsored by the Army Research Laboratory and was
accomplished under the Cooperative Agreement Number W911NF-24-2-0180. The views and conclusions contained in this document are those of the authors and should not be
interpreted as representing the official policies, either expressed or implied, of the Army Research
Laboratory or the U.S. Government.

\section*{Limitations}
Despite our best efforts to develop a comprehensive and robust framework for temporal question answering, several limitations persist.  RASTeR uses \textit{slightly} more resources than traditional prompting. While RASTeR's agent-based architecture introduces multiple prompting steps per query, we found the overall overhead to be manageable in practice. On average, each full query involves 3–4 calls to the underlying LLM, with total token usage averaging 4.64x more that of a single monolithic prompt (Table\ref{tab:cost_appendix}). However, because the number of agent calls is fixed and does not scale with input length or number of retrieved documents, the additional cost remains minimal and predictable across queries. This fixed modular structure ensures stable inference time and simplifies deployment planning. RASTeR has not been evaluated on datasets with gold-standard temporal graphs, leaving the accuracy of its generated knowledge graphs unverified. While the framework is practical in retrieval-based settings, it underperforms on tasks requiring abstract generalization, where simpler prompting strategies may suffice. Moreover, although RASTeR prompts with structured temporal knowledge, it does not yet leverage deeper architectural integration, such as graph neural networks or instruction-tuned models, which may offer more effective handling of complex temporal relationships.

\bibliography{custom}

@inproceedings{petroni-etal-2019-language,
    title = "Language Models as Knowledge Bases?",
    author = {Petroni, Fabio  and
      Rockt{\"a}schel, Tim  and
      Riedel, Sebastian  and
      Lewis, Patrick  and
      Bakhtin, Anton  and
      Wu, Yuxiang  and
      Miller, Alexander},
    editor = "Inui, Kentaro  and
      Jiang, Jing  and
      Ng, Vincent  and
      Wan, Xiaojun",
    booktitle = "Proceedings of the 2019 Conference on Empirical Methods in Natural Language Processing and the 9th International Joint Conference on Natural Language Processing (EMNLP-IJCNLP)",
    month = nov,
    year = "2019",
    address = "Hong Kong, China",
    publisher = "Association for Computational Linguistics",
    url = "https://aclanthology.org/D19-1250/",
    doi = "10.18653/v1/D19-1250",
    pages = "2463--2473",
}

@inproceedings{riezler-maxwell-2005-pitfalls,
    title = "On Some Pitfalls in Automatic Evaluation and Significance Testing for {MT}",
    author = "Riezler, Stefan  and
      Maxwell, John T.",
    editor = "Goldstein, Jade  and
      Lavie, Alon  and
      Lin, Chin-Yew  and
      Voss, Clare",
    booktitle = "Proceedings of the {ACL} Workshop on Intrinsic and Extrinsic Evaluation Measures for Machine Translation and/or Summarization",
    month = jun,
    year = "2005",
    address = "Ann Arbor, Michigan",
    publisher = "Association for Computational Linguistics",
    url = "https://aclanthology.org/W05-0908/",
    pages = "57--64"
}

@article{wang2025lmars,
  title        = {L-MARS: Legal Multi-Agent Workflow with Orchestrated Reasoning and Agentic Search},
  author       = {Wang, Ziqi and Yuan, Boqin},
  journal      = {arXiv preprint arXiv:2509.00761},
  year         = {2025},
  month        = {September},
  url          = {https://doi.org/10.48550/arXiv.2509.00761},
  doi          = {10.48550/arXiv.2509.00761}
}

@inproceedings{chang2025mainrag,
  title        = {MAIN-RAG: Multi-Agent Filtering Retrieval-Augmented Generation},
  author       = {Chang, Chia-Yuan and Jiang, Zhimeng and Rakesh, Vineeth and others},
  booktitle    = {Proceedings of the 63rd Annual Meeting of the Association for Computational Linguistics (Volume 1: Long Papers)},
  editor       = {Che, Wanxiang and Nabende, Joyce and Shutova, Ekaterina and Pilehvar, Mohammad Taher},
  year         = {2025},
  publisher    = {Association for Computational Linguistics},
  url          = {https://doi.org/10.18653/v1/2025.acl-long.131},
  doi          = {10.18653/v1/2025.acl-long.131}
}

@article{nguyen2025marag,
  title        = {MA-RAG: Multi-Agent Retrieval-Augmented Generation via Collaborative Chain-of-Thought Reasoning},
  author       = {Nguyen, Thang and Chin, Peter and Tai, Yu-Wing},
  journal      = {arXiv preprint arXiv:2505.20096},
  year         = {2025},
  month        = {October},
  url          = {https://doi.org/10.48550/arXiv.2505.20096},
  doi          = {10.48550/arXiv.2505.20096}
}

@article{lewis2020retrieval,
  title={Retrieval-augmented generation for knowledge-intensive nlp tasks},
  author={Lewis, Patrick and Perez, Ethan and Piktus, Aleksandra and Petroni, Fabio and Karpukhin, Vladimir and Goyal, Naman and K{\"u}ttler, Heinrich and Lewis, Mike and Yih, Wen-tau and Rockt{\"a}schel, Tim and others},
  journal={Advances in neural information processing systems},
  volume={33},
  pages={9459--9474},
  year={2020}
}

@inproceedings{wallat2024temporal,
  title={Temporal blind spots in large language models},
  author={Wallat, Jonas and Jatowt, Adam and Anand, Avishek},
  booktitle={Proceedings of the 17th ACM International Conference on Web Search and Data Mining},
  pages={683--692},
  year={2024}
}

@article{zhang2025lost,
  title={Lost-in-the-Middle in Long-Text Generation: Synthetic Dataset, Evaluation Framework, and Mitigation},
  author={Zhang, Junhao and Zhang, Richong and Kong, Fanshuang and Miao, Ziyang and Ye, Yanhan and Zheng, Yaowei},
  journal={arXiv preprint arXiv:2503.06868},
  year={2025}
}

@article{liu2024lost,
  title={Lost in the Middle: How Language Models Use Long Contexts},
  author={Liu, Nelson F and Lin, Kevin and Hewitt, John and Paranjape, Ashwin and Bevilacqua, Michele and Petroni, Fabio and Liang, Percy},
  journal={Transactions of the Association for Computational Linguistics},
  volume={12},
  year={2024}
}

@inproceedings{wu2024timesensitive,
  title     = {Time\textendash Sensitive Retrieval\textendash Augmented Generation for Question Answering},
  author    = {Wu, Feifan and Liu, Lingyuan and He, Wentao and Liu, Ziqi and Zhang, Zhiqiang and Wang, Haofen and Wang, Meng},
  booktitle = {Proceedings of the 33rd ACM International Conference on Information and Knowledge Management (CIKM)},
  pages     = {2544--2553},
  year      = {2024},
  doi       = {10.1145/3627673.3679800}
}

@article{chu2023timebench,
  title   = {TimeBench: A Comprehensive Evaluation of Temporal Reasoning Abilities in Large Language Models},
  author  = {Chu, Zheng and Chen, Jingchang and Chen, Qianglong and Yu, Weijiang and Wang, Haotian and Liu, Ming and Qin, Bing},
  journal = {arXiv preprint arXiv:2311.17667},
  year    = {2023}
}

@inproceedings{wang2024tram,
  title     = {TRAM: Benchmarking Temporal Reasoning for Large Language Models},
  author    = {Wang, Yuqing and Zhao, Yun},
  booktitle = {Findings of the Association for Computational Linguistics: ACL 2024},
  year      = {2024},
  doi       = {10.48550/arXiv.2310.00835}
}

@inproceedings{beniwal2024remember,
  title     = {Remember This Event That Year? Assessing Temporal Information and Understanding in Large Language Models},
  author    = {Beniwal, Himanshu and Patel, Dishant and Nandagopan~D, Kowsik and Ladia, Hritik and Yadav, Ankit and Singh, Mayank},
  booktitle = {Findings of the Association for Computational Linguistics: EMNLP 2024},
  pages     = {16239--16348},
  year      = {2024}
}

@article{petroni2020context,
  title={How context affects language models' factual predictions},
  author={Petroni, Fabio and Lewis, Patrick and Piktus, Aleksandra and Rockt{\"a}schel, Tim and Wu, Yuxiang and Miller, Alexander H and Riedel, Sebastian},
  journal={arXiv preprint arXiv:2005.04611},
  year={2020}
}

@inproceedings{sultan2024structured,
  title={Structured Chain-of-Thought Prompting for Few-Shot Generation of Content-Grounded QA Conversations},
  author={Sultan, Md Arafat and Ganhotra, Jatin and Astudillo, Ram{\'o}n Fernandez},
  booktitle={Findings of the Association for Computational Linguistics: EMNLP 2024},
  pages={16172--16187},
  year={2024}
}

@inproceedings{shaier2024desiderata,
  title     = {Desiderata for the Context Use of Question Answering Systems},
  author    = {Shaier, Sagi and Hunter, Lawrence E. and von der Wense, Katharina},
  booktitle = {Proceedings of the 18th Conference of the European Chapter of the Association for Computational Linguistics (EACL)},
  year      = {2024},
  note      = {Long Paper},
  doi       = {10.48550/arXiv.2401.18001},
  url       = {https://arxiv.org/abs/2401.18001}
}

@inproceedings{roberts-etal-2020-much,
    title = "How Much Knowledge Can You Pack Into the Parameters of a Language Model?",
    author = "Roberts, Adam  and
      Raffel, Colin  and
      Shazeer, Noam",
    editor = "Webber, Bonnie  and
      Cohn, Trevor  and
      He, Yulan  and
      Liu, Yang",
    booktitle = "Proceedings of the 2020 Conference on Empirical Methods in Natural Language Processing (EMNLP)",
    month = nov,
    year = "2020",
    address = "Online",
    publisher = "Association for Computational Linguistics",
    url = "https://aclanthology.org/2020.emnlp-main.437/",
    doi = "10.18653/v1/2020.emnlp-main.437",
    pages = "5418--5426",
}

@inproceedings{xiao2021next,
    title = {Next-qa: Next phase of question-answering to explaining temporal actions},
    author = {Xiao, Junbin and Shang, Xindi and Yao, Angela and Chua, Tat-Seng},
    booktitle = {Proceedings of the IEEE/CVF conference on computer vision and pattern recognition},
    year = {2021},
    pages = {9777--9786}
}

@inproceedings{he-etal-2024-retrieving,
    title = "Retrieving, Rethinking and Revising: The Chain-of-Verification Can Improve Retrieval Augmented Generation",
    author = "He, Bolei  and
      Chen, Nuo  and
      He, Xinran  and
      Yan, Lingyong  and
      Wei, Zhenkai  and
      Luo, Jinchang  and
      Ling, Zhen-Hua",
    editor = "Al-Onaizan, Yaser  and
      Bansal, Mohit  and
      Chen, Yun-Nung",
    booktitle = "Findings of the Association for Computational Linguistics: EMNLP 2024",
    month = nov,
    year = "2024",
    address = "Miami, Florida, USA",
    publisher = "Association for Computational Linguistics",
    url = "https://aclanthology.org/2024.findings-emnlp.607/",
    doi = "10.18653/v1/2024.findings-emnlp.607",
    pages = "10371--10393",
}

@inproceedings{in-etal-2025-diversify,
    title = "Diversify-verify-adapt: Efficient and Robust Retrieval-Augmented Ambiguous Question Answering",
    author = "In, Yeonjun  and
      Kim, Sungchul  and
      Rossi, Ryan A.  and
      Tanjim, Mehrab  and
      Yu, Tong  and
      Sinha, Ritwik  and
      Park, Chanyoung",
    editor = "Chiruzzo, Luis  and
      Ritter, Alan  and
      Wang, Lu",
    booktitle = "Proceedings of the 2025 Conference of the Nations of the Americas Chapter of the Association for Computational Linguistics: Human Language Technologies (Volume 1: Long Papers)",
    month = apr,
    year = "2025",
    address = "Albuquerque, New Mexico",
    publisher = "Association for Computational Linguistics",
    url = "https://aclanthology.org/2025.naacl-long.56/",
    doi = "10.18653/v1/2025.naacl-long.56",
    pages = "1212--1233",
    ISBN = "979-8-89176-189-6",
}

@inproceedings{chenghaozhu-etal-2025-llm,
    title = "Is Your {LLM} Outdated? A Deep Look at Temporal Generalization",
    author = "Zhu, Chenghao  and
      Chen, Nuo  and
      Gao, Yufei  and
      Zhang, Yunyi  and
      Tiwari, Prayag  and
      Wang, Benyou",
    editor = "Chiruzzo, Luis  and
      Ritter, Alan  and
      Wang, Lu",
    booktitle = "Proceedings of the 2025 Conference of the Nations of the Americas Chapter of the Association for Computational Linguistics: Human Language Technologies (Volume 1: Long Papers)",
    month = apr,
    year = "2025",
    address = "Albuquerque, New Mexico",
    publisher = "Association for Computational Linguistics",
    url = "https://aclanthology.org/2025.naacl-long.381/",
    doi = "10.18653/v1/2025.naacl-long.381",
    pages = "7433--7457",
    ISBN = "979-8-89176-189-6",
}

@inproceedings{uddin-etal-2025-unseentimeqa,
    title = "{U}n{S}een{T}ime{QA}: Time-Sensitive Question-Answering Beyond {LLM}s' Memorization",
    author = "Uddin, Md Nayem and
              Saeidi, Amir and
              Handa, Divij and
              Seth, Agastya and
              Son, Tran Cao and
              Blanco, Eduardo and
              Corman, Steven and
              Baral, Chitta",
    editor = "Che, Wanxiang and
              Nabende, Joyce and
              Shutova, Ekaterina and
              Pilehvar, Mohammad Taher",
    booktitle = "Proceedings of the 63rd Annual Meeting of the Association for Computational Linguistics (ACL 2025), Volume 1: Long Papers",
    month = jul,
    year = "2025",
    address = "Vienna, Austria",
    publisher = "Association for Computational Linguistics",
    url = "https://aclanthology.org/2025.acl-long.94/",
    pages = "1873--1913",
    isbn = "979-8-89176-251-0"
}

@article{sun2013temporal,
  author       = {Weiyi Sun and Anna Rumshisky and Ozlem Uzuner},
  title        = {Temporal reasoning over clinical text: the state of the art},
  journal      = {Journal of the American Medical Informatics Association},
  volume       = {20},
  number       = {5},
  pages        = {814--819},
  year         = {2013},
  month        = {September},
  doi          = {10.1136/amiajnl-2013-001760},
  url          = {https://doi.org/10.1136/amiajnl-2013-001760}
}

@article{10.1007/s11263-017-1033-7,
author = {Zhu, Linchao and Xu, Zhongwen and Yang, Yi and Hauptmann, Alexander G.},
title = {Uncovering the Temporal Context for Video Question Answering},
year = {2017},
issue_date = {September 2017},
publisher = {Kluwer Academic Publishers},
address = {USA},
volume = {124},
number = {3},
issn = {0920-5691},
url = {https://doi-org.libweb.lib.utsa.edu/10.1007/s11263-017-1033-7},
doi = {10.1007/s11263-017-1033-7},
journal = {Int. J. Comput. Vision},
month = sep,
pages = {409–421},
numpages = {13}
}

@inproceedings{bajpai-etal-2024-temporally,
    title = "Temporally Consistent Factuality Probing for Large Language Models",
    author = "Bajpai, Ashutosh  and
      Goyal, Aaryan  and
      Anwer, Atif  and
      Chakraborty, Tanmoy",
    editor = "Al-Onaizan, Yaser  and
      Bansal, Mohit  and
      Chen, Yun-Nung",
    booktitle = "Proceedings of the 2024 Conference on Empirical Methods in Natural Language Processing",
    month = nov,
    year = "2024",
    address = "Miami, Florida, USA",
    publisher = "Association for Computational Linguistics",
    url = "https://aclanthology.org/2024.emnlp-main.887/",
    doi = "10.18653/v1/2024.emnlp-main.887",
    pages = "15864--15881",
}

@inproceedings{10.1145/3269206.3269247,
author = {Jia, Zhen and Abujabal, Abdalghani and Saha Roy, Rishiraj and Str\"{o}tgen, Jannik and Weikum, Gerhard},
title = {TEQUILA: Temporal Question Answering over Knowledge Bases},
year = {2018},
isbn = {9781450360142},
publisher = {Association for Computing Machinery},
address = {New York, NY, USA},
url = {https://doi-org.libweb.lib.utsa.edu/10.1145/3269206.3269247},
doi = {10.1145/3269206.3269247},
booktitle = {Proceedings of the 27th ACM International Conference on Information and Knowledge Management},
pages = {1807–1810},
numpages = {4},
location = {Torino, Italy},
series = {CIKM '18}
}

@inproceedings{llorens2015semeval,
    title        = {Semeval-2015 task 5: Qa tempeval-evaluating temporal information understanding with question answering},
    author       = {Llorens, Hector and Chambers, Nathanael and UzZaman, Naushad and Mostafazadeh, Nasrin and Allen, James and Pustejovsky, James},
    year         = 2015,
    booktitle    = {proceedings of the 9th International Workshop on Semantic Evaluation (SemEval 2015)},
    pages        = {792--800}
}

@inproceedings{jang2017tgif,
    title = {Tgif-qa: Toward spatio-temporal reasoning in visual question answering},
    author = {Jang, Yunseok and Song, Yale and Yu, Youngjae and Kim, Youngjin and Kim, Gunhee},
    booktitle = {Proceedings of the IEEE conference on computer vision and pattern recognition},
    year = {2017},
    pages = {2758--2766}
}

@misc{zhang2024knowgptknowledgegraphbased,
      title={KnowGPT: Knowledge Graph based Prompting for Large Language Models}, 
      author={Qinggang Zhang and Junnan Dong and Hao Chen and Daochen Zha and Zailiang Yu and Xiao Huang},
      year={2024},
      eprint={2312.06185},
      archivePrefix={arXiv},
      primaryClass={cs.CL},
      url={https://arxiv.org/abs/2312.06185}, 
}

@inproceedings{wei-etal-2023-menatqa,
    title = "{M}enat{QA}: A New Dataset for Testing the Temporal Comprehension and Reasoning Abilities of Large Language Models",
    author = "Wei, Yifan  and
      Su, Yisong  and
      Ma, Huanhuan  and
      Yu, Xiaoyan  and
      Lei, Fangyu  and
      Zhang, Yuanzhe  and
      Zhao, Jun  and
      Liu, Kang",
    editor = "Bouamor, Houda  and
      Pino, Juan  and
      Bali, Kalika",
    booktitle = "Findings of the Association for Computational Linguistics: EMNLP 2023",
    month = dec,
    year = "2023",
    address = "Singapore",
    publisher = "Association for Computational Linguistics",
    url = "https://aclanthology.org/2023.findings-emnlp.100/",
    doi = "10.18653/v1/2023.findings-emnlp.100",
    pages = "1434--1447",
}

@inproceedings{tan-etal-2023-towards,
    title = "Towards Benchmarking and Improving the Temporal Reasoning Capability of Large Language Models",
    author = "Tan, Qingyu  and
      Ng, Hwee Tou  and
      Bing, Lidong",
    editor = "Rogers, Anna  and
      Boyd-Graber, Jordan  and
      Okazaki, Naoaki",
    booktitle = "Proceedings of the 61st Annual Meeting of the Association for Computational Linguistics (Volume 1: Long Papers)",
    month = jul,
    year = "2023",
    address = "Toronto, Canada",
    publisher = "Association for Computational Linguistics",
    url = "https://aclanthology.org/2023.acl-long.828/",
    doi = "10.18653/v1/2023.acl-long.828",
    pages = "14820--14835",
}

@inproceedings{TimeSensitiveQA,
 author = {Chen, Wenhu and Wang, Xinyi and Wang, William Yang and Wang, William Yang},
 booktitle = {Proceedings of the Neural Information Processing Systems Track on Datasets and Benchmarks},
 editor = {J. Vanschoren and S. Yeung},
 pages = {},
 title = {A Dataset for Answering Time-Sensitive Questions},
 url = {https://datasets-benchmarks-proceedings.neurips.cc/paper_files/paper/2021/file/1f0e3dad99908345f7439f8ffabdffc4-Paper-round2.pdf},
 volume = {1},
 year = {2021}
}

@inproceedings{xia-etal-2022-metatkg,
    title = "{M}eta{TKG}: Learning Evolutionary Meta-Knowledge for Temporal Knowledge Graph Reasoning",
    author = "Xia, Yuwei  and
      Zhang, Mengqi  and
      Liu, Qiang  and
      Wu, Shu  and
      Zhang, Xiao-Yu",
    editor = "Goldberg, Yoav  and
      Kozareva, Zornitsa  and
      Zhang, Yue",
    booktitle = "Proceedings of the 2022 Conference on Empirical Methods in Natural Language Processing",
    month = dec,
    year = "2022",
    address = "Abu Dhabi, United Arab Emirates",
    publisher = "Association for Computational Linguistics",
    url = "https://aclanthology.org/2022.emnlp-main.487/",
    doi = "10.18653/v1/2022.emnlp-main.487",
    pages = "7230--7240",
   }

@inproceedings{chen-etal-2024-unified,
    title = "A Unified Temporal Knowledge Graph Reasoning Model Towards Interpolation and Extrapolation",
    author = "Chen, Kai  and
      Wang, Ye  and
      Li, Yitong  and
      Li, Aiping  and
      Yu, Han  and
      Song, Xin",
    editor = "Ku, Lun-Wei  and
      Martins, Andre  and
      Srikumar, Vivek",
    booktitle = "Proceedings of the 62nd Annual Meeting of the Association for Computational Linguistics (Volume 1: Long Papers)",
    month = aug,
    year = "2024",
    address = "Bangkok, Thailand",
    publisher = "Association for Computational Linguistics",
    url = "https://aclanthology.org/2024.acl-long.8/",
    doi = "10.18653/v1/2024.acl-long.8",
    pages = "117--132",
}

@misc{gao2024twostagegenerativequestionanswering,
      title={Two-stage Generative Question Answering on Temporal Knowledge Graph Using Large Language Models}, 
      author={Yifu Gao and Linbo Qiao and Zhigang Kan and Zhihua Wen and Yongquan He and Dongsheng Li},
      year={2024},
      eprint={2402.16568},
      archivePrefix={arXiv},
      primaryClass={cs.CL},
      url={https://arxiv.org/abs/2402.16568}, 
}

@inproceedings{qianyihu-etal-2025-time,
    title = "Time-aware {R}e{A}ct Agent for Temporal Knowledge Graph Question Answering",
    author = "Hu, Qianyi  and
      Tu, Xinhui  and
      Guo, Cong  and
      Zhang, Shunping",
    editor = "Chiruzzo, Luis  and
      Ritter, Alan  and
      Wang, Lu",
    booktitle = "Findings of the Association for Computational Linguistics: NAACL 2025",
    month = apr,
    year = "2025",
    address = "Albuquerque, New Mexico",
    publisher = "Association for Computational Linguistics",
    url = "https://aclanthology.org/2025.findings-naacl.334/",
    doi = "10.18653/v1/2025.findings-naacl.334",
    pages = "6013--6024",
    ISBN = "979-8-89176-195-7",
}

@misc{gruber2024complextempqalargescaledatasetcomplex,
      title={ComplexTempQA: A Large-Scale Dataset for Complex Temporal Question Answering}, 
      author={Raphael Gruber and Abdelrahman Abdallah and Michael Färber and Adam Jatowt},
      year={2024},
      eprint={2406.04866},
      archivePrefix={arXiv},
      primaryClass={cs.CL},
      url={https://arxiv.org/abs/2406.04866}, 
}

@inproceedings{velupillai-etal-2015-blulab,
    title = "{B}lu{L}ab: Temporal Information Extraction for the 2015 Clinical {T}emp{E}val Challenge",
    author = "Velupillai, Sumithra  and
      Mowery, Danielle L  and
      Abdelrahman, Samir  and
      Christensen, Lee  and
      Chapman, Wendy",
    editor = "Nakov, Preslav  and
      Zesch, Torsten  and
      Cer, Daniel  and
      Jurgens, David",
    booktitle = "Proceedings of the 9th International Workshop on Semantic Evaluation ({S}em{E}val 2015)",
    month = jun,
    year = "2015",
    address = "Denver, Colorado",
    publisher = "Association for Computational Linguistics",
    url = "https://aclanthology.org/S15-2137/",
    doi = "10.18653/v1/S15-2137",
    pages = "815--819"
}

@inproceedings{
wu2025chainoftimeline,
title={Chain-of-Timeline: Enhancing {LLM} Zero-Shot Temporal Reasoning with {SQL}-Style Timeline Formalization},
author={Jiaying Wu and Bryan Hooi},
booktitle={Workshop on Reasoning and Planning for Large Language Models},
year={2025},
url={https://openreview.net/forum?id=ptHtrHwMpw}
}

@inproceedings{saxena2021question,
    title = {Question Answering Over Temporal Knowledge Graphs},
    author = {Saxena, Apoorv and Chakrabarti, Soumen and Talukdar, Partha},
    booktitle = {Proceedings of the 59th Annual Meeting of the Association for Computational Linguistics and the 11th International Joint Conference on Natural Language Processing (Volume 1: Long Papers)},
    year = {2021},
    pages = {6663--6676}
}

@inproceedings{zhao2024utsa,
    title = {UTSA-NLP at ChemoTimelines 2024: Evaluating Instruction-Tuned Language Models for Temporal Relation Extraction},
    author = {Zhao, Xingmeng and Rios, Anthony},
    booktitle = {Proceedings of the 6th Clinical Natural Language Processing Workshop},
    year = {2024}
}

@misc{han2025retrievalaugmentedgenerationgraphsgraphrag,
      title={Retrieval-Augmented Generation with Graphs (GraphRAG)}, 
      author={Haoyu Han and Yu Wang and Harry Shomer and Kai Guo and Jiayuan Ding and Yongjia Lei and Mahantesh Halappanavar and Ryan A. Rossi and Subhabrata Mukherjee and Xianfeng Tang and Qi He and Zhigang Hua and Bo Long and Tong Zhao and Neil Shah and Amin Javari and Yinglong Xia and Jiliang Tang},
      year={2025},
      eprint={2501.00309},
      archivePrefix={arXiv},
      primaryClass={cs.IR},
      url={https://arxiv.org/abs/2501.00309}, 
}

@inproceedings{wang2022archivalqa,
    title = {ArchivalQA: a large-scale benchmark dataset for open-domain question answering over historical news collections},
    author = {Wang, Jiexin and Jatowt, Adam and Yoshikawa, Masatoshi},
    booktitle = {Proceedings of the 45th International ACM SIGIR Conference on Research and Development in Information Retrieval},
    year = {2022},
    pages = {3025--3035}
}

@inproceedings{kim-etal-2024-aligning,
    title = "Aligning Language Models to Explicitly Handle Ambiguity",
    author = "Kim, Hyuhng Joon  and
      Kim, Youna  and
      Park, Cheonbok  and
      Kim, Junyeob  and
      Park, Choonghyun  and
      Yoo, Kang Min  and
      Lee, Sang-goo  and
      Kim, Taeuk",
    editor = "Al-Onaizan, Yaser  and
      Bansal, Mohit  and
      Chen, Yun-Nung",
    booktitle = "Proceedings of the 2024 Conference on Empirical Methods in Natural Language Processing",
    month = nov,
    year = "2024",
    address = "Miami, Florida, USA",
    publisher = "Association for Computational Linguistics",
    url = "https://aclanthology.org/2024.emnlp-main.119/",
    doi = "10.18653/v1/2024.emnlp-main.119",
    pages = "1989--2007",
}

@misc{gruber2024complextempqa,
    title = {ComplexTempQA: A Large-Scale Dataset for Complex Temporal Question Answering},
    author = {Gruber, Raphael and Abdallah, Abdelrahman and Färber, Michael and Jatowt, Adam},
    year = {2024},
    eprint = {2406.04866},
    archiveprefix = {arXiv},
    primaryclass = {cs.CL}
}

@misc{hu2025mctsragenhancingretrievalaugmentedgeneration,
      title={MCTS-RAG: Enhancing Retrieval-Augmented Generation with Monte Carlo Tree Search}, 
      author={Yunhai Hu and Yilun Zhao and Chen Zhao and Arman Cohan},
      year={2025},
      eprint={2503.20757},
      archivePrefix={arXiv},
      primaryClass={cs.CL},
      url={https://arxiv.org/abs/2503.20757}, 
}

@article{qiu2023large,
    title = {Are Large Language Models Temporally Grounded?},
    author = {Qiu, Yifu and Zhao, Zheng and Ziser, Yftah and Korhonen, Anna and Ponti, Edoardo M and Cohen, Shay B},
    journal = {arXiv preprint arXiv:2311.08398},
    year = {2023}
}

@article{tan2023towards,
    title = {Towards benchmarking and improving the temporal reasoning capability of large language models},
    author = {Tan, Qingyu and Ng, Hwee Tou and Bing, Lidong},
    journal = {arXiv preprint arXiv:2306.08952},
    year = {2023}
}

@article{gao2024two,
    title = {Two-stage Generative Question Answering on Temporal Knowledge Graph Using Large Language Models},
    author = {Gao, Yifu and Qiao, Linbo and Kan, Zhigang and Wen, Zhihua and He, Yongquan and Li, Dongsheng},
    journal = {arXiv preprint arXiv:2402.16568},
    year = {2024}
}

@inproceedings{cheng2024exploring,
    title = {Exploring the Robustness of In-Context Learning with Noisy Labels},
    author = {Chen Cheng and Xinzhi Yu and Haodong Wen and Jingsong Sun and Guanzhang Yue and Yihao Zhang and Zeming Wei},
    booktitle = {ICLR 2024 Workshop on Reliable and Responsible Foundation Models},
    year = {2024},
    url = {https://openreview.net/forum?id=ib4cAWZKXa}
}

@inproceedings{yoran2024retrieval,
    title = {Making Retrieval-Augmented Language Models Robust to Irrelevant Context},
    author = {Yoran, Ori and Wolfson, Tomer and Ram, Ori and Berant, Jonathan},
    booktitle = {International Conference on Learning Representations (ICLR)},
    year = {2024}
}

@misc{wang2025retrievalaugmentedgenerationconflictingevidence,
      title={Retrieval-Augmented Generation with Conflicting Evidence}, 
      author={Han Wang and Archiki Prasad and Elias Stengel-Eskin and Mohit Bansal},
      year={2025},
      eprint={2504.13079},
      archivePrefix={arXiv},
      primaryClass={cs.CL},
      url={https://arxiv.org/abs/2504.13079}, 
}

@misc{yoran2024makingretrievalaugmentedlanguagemodels,
      title={Making Retrieval-Augmented Language Models Robust to Irrelevant Context}, 
      author={Ori Yoran and Tomer Wolfson and Ori Ram and Jonathan Berant},
      year={2024},
      eprint={2310.01558},
      archivePrefix={arXiv},
      primaryClass={cs.CL},
      url={https://arxiv.org/abs/2310.01558}, 
}

@misc{fang2024enhancingnoiserobustnessretrievalaugmented,
      title={Enhancing Noise Robustness of Retrieval-Augmented Language Models with Adaptive Adversarial Training}, 
      author={Feiteng Fang and Yuelin Bai and Shiwen Ni and Min Yang and Xiaojun Chen and Ruifeng Xu},
      year={2024},
      eprint={2405.20978},
      archivePrefix={arXiv},
      primaryClass={cs.AI},
      url={https://arxiv.org/abs/2405.20978}, 
}

@inproceedings{wallat-etal-2025-study,
    title = "A Study into Investigating Temporal Robustness of {LLM}s",
    author = "Wallat, Jonas  and
      Abdallah, Abdelrahman  and
      Jatowt, Adam  and
      Anand, Avishek",
    editor = "Che, Wanxiang  and
      Nabende, Joyce  and
      Shutova, Ekaterina  and
      Pilehvar, Mohammad Taher",
    booktitle = "Findings of the Association for Computational Linguistics: ACL 2025",
    month = jul,
    year = "2025",
    address = "Vienna, Austria",
    publisher = "Association for Computational Linguistics",
    url = "https://aclanthology.org/2025.findings-acl.810/",
    pages = "15685--15705",
    ISBN = "979-8-89176-256-5",
}

@inproceedings{luu-etal-2022-time,
    title = "Time Waits for No One! Analysis and Challenges of Temporal Misalignment",
    author = "Luu, Kelvin  and
      Khashabi, Daniel  and
      Gururangan, Suchin  and
      Mandyam, Karishma  and
      Smith, Noah A.",
    editor = "Carpuat, Marine  and
      de Marneffe, Marie-Catherine  and
      Meza Ruiz, Ivan Vladimir",
    booktitle = "Proceedings of the 2022 Conference of the North American Chapter of the Association for Computational Linguistics: Human Language Technologies",
    month = jul,
    year = "2022",
    address = "Seattle, United States",
    publisher = "Association for Computational Linguistics",
    url = "https://aclanthology.org/2022.naacl-main.435/",
    doi = "10.18653/v1/2022.naacl-main.435",
    pages = "5944--5958",
}

@inproceedings{tan-etal-2024-towards,
    title = "Towards Robust Temporal Reasoning of Large Language Models via a Multi-Hop {QA} Dataset and Pseudo-Instruction Tuning",
    author = "Tan, Qingyu  and
      Ng, Hwee Tou  and
      Bing, Lidong",
    editor = "Ku, Lun-Wei  and
      Martins, Andre  and
      Srikumar, Vivek",
    booktitle = "Findings of the Association for Computational Linguistics: ACL 2024",
    month = aug,
    year = "2024",
    address = "Bangkok, Thailand",
    publisher = "Association for Computational Linguistics",
    url = "https://aclanthology.org/2024.findings-acl.374/",
    doi = "10.18653/v1/2024.findings-acl.374",
    pages = "6272--6286",
}

@inproceedings{yin-etal-2023-alcuna,
    title = "{ALCUNA}: Large Language Models Meet New Knowledge",
    author = "Yin, Xunjian  and
      Huang, Baizhou  and
      Wan, Xiaojun",
    editor = "Bouamor, Houda  and
      Pino, Juan  and
      Bali, Kalika",
    booktitle = "Proceedings of the 2023 Conference on Empirical Methods in Natural Language Processing",
    month = dec,
    year = "2023",    address = "Singapore",
    publisher = "Association for Computational Linguistics",
    url = "https://aclanthology.org/2023.emnlp-main.87/",
    doi = "10.18653/v1/2023.emnlp-main.87",
    pages = "1397--1414",
}

@article{Lorenzini2022,
  title={Protest Event Analysis: Developing a Semiautomated NLP Approach},
  author={Jasmine Lorenzini and Hanspeter Kriesi and Peter Makarov and Bruno W{\"u}est},
  journal={American Behavioral Scientist},
  volume={66},
  number={5},
  pages={555--577},
  year={2022},
  publisher={SAGE Publications},
  doi={10.1177/00027642211021650},
  url={https://doi.org/10.1177/00027642211021650},
  license={CC BY 4.0}
}

@misc{park2024enhancingrobustnessretrievalaugmentedlanguage,
      title={Enhancing Robustness of Retrieval-Augmented Language Models with In-Context Learning}, 
      author={Seong-Il Park and Seung-Woo Choi and Na-Hyun Kim and Jay-Yoon Lee},
      year={2024},
      eprint={2408.04414},
      archivePrefix={arXiv},
      primaryClass={cs.CL},
      url={https://arxiv.org/abs/2408.04414}, 
}

\appendix
\section{Appendix}
\label{sec:appendix}
\subsection{Metric Formalization}
\label{appendix:metrics}
In any NLP applications, due to the diverse nature of natural langue, determining the correctness of a prediction is always challenging. To highlight this challenge, Figure~\ref{fig:em_failure_example} shows how the model output can be marked incorrect by both exact match (EM) and contains accuracy (Acc), despite being semantically correct. Having a variety of evaluation metrics allows us to get a better picture of model performance measured by partial matches, and more strict criteria.

\begin{figure}[h]
\centering
\begin{tcolorbox}[colback=gray!5, colframe=gray!75!black, width=0.9\linewidth, sharp corners=south]
\textbf{QUESTION:} John O. Moseley was an employee for whom from Mar 1936 to Dec 1938? \\
\textbf{OUTPUT:} central state college \\
\textbf{GROUND TRUTH:} central state teachers college
\end{tcolorbox}
\caption{An example where the model output is semantically correct but fails EM and Acc.}
\label{fig:em_failure_example}
\end{figure}

To define our evaluation metrics formally, Let $\hat{a}$ be the predicted answer and let $A = \{a_1, a_2, \dots, a_n\}$ denote the set of gold reference answers. Let $W_x$ represent the multiset of words in answer $x$.

\vspace{2mm} \noindent \textbf{Exact Match (EM).}  
EM measures whether the ground truth is \textit{exactly identical} to the prediction. (e.g. "Border Collie" is identical to "Border Collie")
\[
\text{EM} = \mathbf{1}[\hat{a} \in A]
\]  
EM returns 1 if the predicted answer exactly matches any gold answer, and 0 otherwise.

\vspace{2mm} \noindent \textbf{Contains Accuracy (Acc).}  
Acc measures whether the ground truth \textit{is contained} in the prediction. (e.g. "Border Collie" is contained in "The dog is a Border Collie")
\[
\text{Acc} = \mathbf{1}[\exists a \in A \text{ such that } a \subseteq \hat{a}]
\]  
Acc returns 1 if any gold answer is a substring of the predicted answer, and 0 otherwise.

\vspace{2mm} \noindent \textbf{Word-Level F1.}  
F1 is the most flexible metric. It measures the maximum word overlap between the predicted and gold answers by computing the harmonic mean of precision and recall. For each $a \in A$, we compute:

\[
\text{Precision} = \frac{|W_{\hat{a}} \cap W_a|}{|W_{\hat{a}}|}, \quad
\text{Recall} = \frac{|W_{\hat{a}} \cap W_a|}{|W_a|}
\]
\[
\text{F1} = \max_{a \in A} \frac{2 \cdot \text{Precision} \cdot \text{Recall}}{\text{Precision} + \text{Recall}}
\]

For example: if the predicted answer is "central state college" and the gold answer is "central state teachers college", the prediction shares three words with the gold answer. Precision is $1$ (3 out of 3 words), recall is $.75$ (3 out of 4 words), and F1 = $\frac{2 \cdot 1 \cdot .75}{1 + .75} = .857$.

\subsection{Descriptive Statistics}
\label{appendix:Descriptive Statistics}
In Table~\ref{tab:discriptive_statistics}, we report the average number of words per context and the number of samples ($n$) for all datasets used in our experiments. The full MQA dataset was included but is substantially smaller than the other datasets. Full subsets of UTQA were also used, though we excluded the easy and medium settings, as they were less challenging and required minimal reasoning compared to the hard subsets. Among all subsets, HS n\_20 had the highest average word count, with nearly 5.5k words. This is due to the relevant context being surrounded by forty distractors. The TSQA subsets also had long contexts, making them the second most verbose in terms of average word count.

\begin{table}[h]
\centering
\resizebox{\linewidth}{!}{
\begin{tabular}{llll}
\toprule
\textbf{Dataset} & \textbf{Subset} & \textbf{Avg. Words} & \textbf{n} \\
\midrule
MQA        & counterfactual & 82.38   & 112  \\
MQA        & order          & 80.44   & 182  \\
MQA        & scope          & 82.38   & 112  \\
MQA        & scope\_expand  & 75.91   & 176  \\
UTQA   & hardSerial         & 140.84  & 2700 \\
UTQA   & hardParallel       & 140.47  & 2700 \\
HS       & n\_1             & 399.94  & 200  \\
HS       & n\_3             & 904.67  & 200  \\
HS       & n\_5             & 1503.09 & 200  \\
HS       & n\_7           & 2021.04 & 200  \\
HS       & n\_20          & 5494.45 & 200  \\
TR     & l2             & 128.29  & 250  \\
TR     & l3             & 141.08  & 250  \\
TSQA& easy           & 2041.00 & 250  \\
TSQA         & hard           & 1827.08 & 250  \\
\bottomrule
\end{tabular}}
\caption{Average word count of relevant context and number of samples (n) per subset.}
\label{tab:discriptive_statistics}
\end{table}

For ablations, we used a subset of 80 randomly selected rows sampled from our three main datasets. Half of the rows came from TR, while the other half were drawn from MQA and TSQA. Table~\ref{tab:ablation_descriptive_statistics} summarizes the row counts and proportions for each subset included in the ablation.

\begin{table}[h]
\centering
\scriptsize
\begin{tabular}{lccc}
\toprule
\textbf{Dataset} & \textbf{Subset} & \textbf{Count} & \textbf{Proportion (\%)} \\
\midrule
\multirow{4}{*}{MQA}& counterfactual & 7  & 8.8 \\
  & order          & 6  & 7.5 \\
  & scope          & 5  & 6.2 \\
  & scope\_expand   & 7  & 8.8 \\
\midrule
\multirow{2}{*}{TR}& l2             & 16 & 20.0 \\
  & l3             & 16 & 20.0 \\
\midrule
\multirow{2}{*}{TSQA}& easy           & 14 & 17.5 \\
  & hard           & 9  & 11.2 \\
\midrule
\textbf{Total} & -- & \textbf{80} & \textbf{100.0} \\
\bottomrule
\end{tabular}
\caption{Descriptive statistics of combined data subsets used for ablations}
\label{tab:ablation_descriptive_statistics}
\end{table}

\subsection{Estimated Token Costs}
\label{appendix:Estimated_Token_Costs}
Table~\ref{tab:cost_appendix} reports estimated token usage (in millions) for TKG generation versus our multi-agent RASTeR method across all datasets. Although RASTeR uses more tokens due to its modular agents, the overall increase is modest relative to the scale of improvement in accuracy and robustness.

More often than not, multi-agent frameworks  increase computational cost, as  noted in our limitations section. However, this increase is not a fundamental weakness but a general property of agentic systems. As inference costs continue to decline, we expect this overhead to become increasingly negligible. Even with a 4.64$\times$ average increase  (shown in Table \ref{tab:cost_appendix} ), the total expense remains within a practical range for modern research pipelines.

\begin{table}[t]
\centering
\small
\setlength{\tabcolsep}{8pt}
\renewcommand{\arraystretch}{1.1}
\begin{tabular}{lcc}
\toprule
\textbf{Dataset} & \textbf{TKG} & \textbf{RASTeR} \\
\midrule
MenatQA      & 4.52  & 16.37 \\
TempReason   & 5.14  & 20.20 \\
TimeQA       & 5.75  & 48.91 \\
\midrule
\textbf{Average} & \textbf{4.83} & \textbf{22.43} \\
\bottomrule
\end{tabular}
\caption{Estimated cost in USD per million tokans comparing single-agent TKG generation and the multi-agent RASTeR framework. Despite the multi-agent structure, the relative increase in cost remains moderate.}
\label{tab:cost_appendix}
\end{table}

\subsection{Expanded Results}
In addition to Acc show in Table~\ref{tab:cont_acc_avg}, we report EM in Table~\ref{tab:em_avg} and F1 scores in Table~\ref{tab:f1_avg}. While Gemma with the Few-Shot prompt slightly outperforms RASTeR in terms of EM (by 0.8\%), RASTeR consistently performs better on both Acc and F1. In fact, RASTeR shows the strongest gains when evaluated through the lens of F1. For example, RASTeR combined with LLaMA achieves a full 7\% improvement over the next-best average F1 score.

We believe RASTeR’s strong performance under F1 is due to the metric’s sensitivity to partial overlap. Predictions are often semantically correct but do not match the gold answer word-for-word, especially when context is missing. Since RASTeR is designed to filter, correct, and reason over noisy context, it excels in settings where exact matches are unlikely but partial correctness is common.

\label{appendix:Expanded Results}
\begin{table}[t]
\centering
\resizebox{\linewidth}{!}{
\begin{tabular}{llcccc}
\toprule
\textbf{Model} & \textbf{Prompt Type} & \textbf{MQA} & \textbf{TR} & \textbf{TSQA} & \textbf{Avg} \\
\midrule
\multirow{4}{*}{\textbf{gemma-3-12b-it}} 
& Few-Shot   &     \textbf{.306} & .257 & \textbf{.150} & \textbf{.272} \\
& Reasoning  &     .191 & .266 & .136 & .194 \\
\cmidrule(lr){2-6}
& TKG        &     .291 & .240 & .142 & .258 \\
& RASTeR     &     .291 & \textbf{.276} & .140 & .264 \\
\midrule
\multirow{4}{*}{\textbf{gpt-4o-mini}} 
& Few-Shot   &     .268 & .288 & .188 & .258 \\
& Reasoning  &     .210 & \textbf{.318} & .195 & .226 \\
\cmidrule(lr){2-6}
& TKG        &     .273 & .253 & .177 & .254 \\
& RASTeR     &     \textbf{.287} & .303 & \textbf{.221} & \textbf{.287} \\
\midrule
\multirow{4}{*}{\textbf{Llama-3.1-8B-Instruct}} 
& Few-Shot   &     .056 & .122 & .060 & .068 \\
& Reasoning  &     .088 & .193 & .095 & .107 \\
\cmidrule(lr){2-6}
& TKG        &     .178 & .153 & .093 & .159 \\
& RASTeR     &     \textbf{.213} & \textbf{.222} & \textbf{.148} & \textbf{.204} \\
\bottomrule
\end{tabular}}
\caption{Exact Match (EM) averaged across subset, and eval-context for each model and prompting strategy.}
\label{tab:em_avg}
\end{table}

\begin{table}[t]
\centering
\resizebox{\linewidth}{!}{
\begin{tabular}{llcccc}
\toprule
\textbf{Model} & \textbf{Prompt Type} & \textbf{MQA} & \textbf{TR} & \textbf{TSQA} & \textbf{Avg} \\
\midrule
\multirow{4}{*}{\textbf{gemma-3-12b-it}} 
& Few-Shot   &     .364 & .321 & .206 & .331 \\
& Reasoning  &     .248 & .317 & .211 & .253 \\
\cmidrule(lr){2-6}
& TKG        &     .345 & .304 & .205 & .315 \\
& RASTeR     &     \textbf{.368} & \textbf{.383} & \textbf{.223} & \textbf{.346} \\
\midrule
\multirow{4}{*}{\textbf{gpt-4o-mini}} 
& Few-Shot   &     .343 & .338 & .270 & .330 \\
& Reasoning  &     .292 & .382 & .285 & .306 \\
\cmidrule(lr){2-6}
& TKG        &     .344 & .306 & .248 & .322 \\
& RASTeR     &     \textbf{.359} & \textbf{.380} & \textbf{.317} & \textbf{.364} \\
\midrule
\multirow{4}{*}{\textbf{Llama-3.1-8B-Instruct}} 
& Few-Shot   &     .085 & .171 & .090 & .100 \\
& Reasoning  &     .129 & .239 & .135 & .148 \\
\cmidrule(lr){2-6}
& TKG        &     .225 & .208 & .133 & .207 \\
& RASTeR     &     \textbf{.285} & \textbf{.300} & \textbf{.224} & \textbf{.277} \\
\bottomrule
\end{tabular}}
\caption{F1 Score averaged across subset, and eval-context for each model and prompting strategy.}
\label{tab:f1_avg}
\end{table}

\subsection{Significance Testing}
\label{appendix:significance_testing}
To assess whether the observed performance differences between RASTeR and each baseline model were statistically reliable, we conducted a two-stage significance analysis covering all models, datasets, and evaluation contexts.

\vspace{2mm} \noindent \textbf{Stage 1 – Pairwise Bootstrap Testing.}
For every combination of model, dataset, subset, and evaluation context, we compared RASTeR to each baseline (\textit{generic-fs}, \textit{reasoning-fs}, \textit{TKG-fs}) using a paired bootstrap test on per-example correctness scores (\(1 = \text{correct}, 0 = \text{incorrect}\)).  
Each bootstrap sample resampled the same set of test instances with replacement to estimate the distribution of the accuracy difference:
\[
\Delta = \bar{x}_{\text{RASTeR}} - \bar{x}_{\text{baseline}}.
\]
From 10{,}000 bootstrap replicates, we computed percentile-based confidence intervals and one-sided \(p\)-values following \citet{riezler-maxwell-2005-pitfalls}. The full grid of comparisons consisted of 216 pairwise tests

\vspace{2mm} \noindent \textbf{Stage 2 – Pooled McNemar Tests.}
To summarize effects at the model level, we aggregated contingency counts across all datasets, subsets, and evaluation contexts.  
For each model–baseline pair, we accumulated:
\[
\begin{aligned}
a &= \text{both correct}, \\
b &= \text{RASTeR only correct}, \\
c &= \text{baseline only correct}, \\
d &= \text{both incorrect}.
\end{aligned}
\]
We then applied a continuity-corrected McNemar chi-square test:
\[
\chi^2 = \frac{(|b - c| - 1)^2}{b + c},
\]
which tests whether the two systems differ significantly in accuracy while accounting for paired dependence.
Two-sided \(p\)-values were derived from the chi-square distribution with one degree of freedom.

Across all 216 pairwise tests, every statistically significant difference favored RASTeR (positive \(\Delta\)).  
The pooled McNemar results, summarized in Table \ref{tab:pooled_sig}, confirm that RASTeR’s improvements are robust and consistent across models.

\begin{table}[t]
\centering
\resizebox{\linewidth}{!}{
\begin{tabular}{llcc}
\toprule
\textbf{Model} & \textbf{Baseline} & \textbf{\(p_{\text{McNemar}}\)} & \textbf{Significant (0.05)} \\
\midrule
\multirow{3}{*}{\textbf{gemma-3-12b-it}} 
 & TKG\_fs       & $3.8 \times 10^{-5}$   & Yes \\
 & Generic\_fs   & 0.132                  & No  \\
 & Reasoning\_fs & $2.2 \times 10^{-11}$  & Yes \\
\midrule
\multirow{3}{*}{\textbf{gpt-4o-mini}} 
 & TKG\_fs       & $1.4 \times 10^{-17}$  & Yes \\
 & Generic\_fs   & $2.6 \times 10^{-9}$   & Yes \\
 & Reasoning\_fs & $1.3 \times 10^{-9}$   & Yes \\
\midrule
\multirow{3}{*}{\textbf{LLaMA-3.1-8B-Instruct}} 
 & TKG\_fs       & 0.0596                 & No  \\
 & Generic\_fs   & $6.5 \times 10^{-115}$ & Yes \\
 & Reasoning\_fs & $1.1 \times 10^{-7}$   & Yes \\
\bottomrule
\end{tabular}}
\caption{Pooled McNemar significance tests comparing RASTeR against each baseline, aggregated across all datasets, subsets, and evaluation contexts. Significant results (\(p<0.05\)) indicate that RASTeR’s accuracy improvements are statistically reliable.}
\label{tab:pooled_sig}
\end{table}

\subsection{Using Semantically Similar Context}
\label{appendix:semantically_similar_context}

To further ensure that our system is robust against realistic confounders, we introduce a \textbf{semantically similar} context, denoted \(C_s\).

\(C_s\) is selected based on cosine similarity between the question embedding and other context embeddings, computed using the \texttt{all-MiniLM-L6-v2} model. Formally,

{\scriptsize
\[
C_s^{(n)} = C_r^{(m)}, \quad 
m = \arg\max_{j \in \{1,\dots,N\} \setminus \{n\}}
\cos\!\big(\mathrm{emb}(Q_n),\, \mathrm{emb}(C_r^{(j)})\big)
\]
}

This procedure selects the most semantically similar yet distinct context to \(Q_n\), allowing us to evaluate model robustness under \textit{plausible but misleading} evidence rather than purely random noise~\footnote{While questions within each dataset share similar themes, introducing semantically similar distractors provides an additional test of the system’s stability under realistic retrieval conditions.}. 

In the haystack experiments, distractor passages were previously selected at random. Likewise, the irrelevant-context experiments also relied on random sampling. To test the robustness of our approach under more realistic retrieval noise, we conducted additional experiments using \textit{semantically similar} contexts, identified using cosine similarity between question and context embeddings computed with \texttt{all-MiniLM-L6-v2}. These experiments help determine whether RASTeR remains effective when distractors are not arbitrary but instead closely related to the query in meaning.

\vspace{2mm} \noindent \textbf{Main Experiment.}
Table~\ref{tab:sem_sim_irrelevant} extends the results reported in Table~2 by evaluating the \texttt{gpt-4o-mini} model on the \textsc{TempReason} dataset under semantically similar distractors.

\begin{table}[h!]
\centering
\small
\begin{tabular}{lcc}
\toprule
\textbf{Prompt Type} & \textbf{Dataset} & \textbf{Cont. Acc.} \\
\midrule
Generic Few-shot & TempReason & 0.045 \\
Reasoning Few-shot & TempReason & 0.135 \\
RASTeR & TempReason & \textbf{0.153} \\
\bottomrule
\end{tabular}
\caption{gpt-4o-mini's performance on  semantically similar irrelevant contexts }
\label{tab:sem_sim_irrelevant}
\end{table}

As shown, RASTeR continues to outperform both generic and reasoning few-shot baselines even when distractors are not chosen at random but by semantics. 

\vspace{2mm} \noindent \textbf{Needle-in-a-Haystack Experiment.}
We also extended the needle-in-a-haystack experiments to use semantically similar distractors. The results with \texttt{gpt-4o-mini} are presented in Table~\ref{tab:sem_sim_haystack}.

\begin{table}[h!]
\centering
\small
\begin{tabular}{lccc}
\toprule
\textbf{Prompt Type} & \textbf{Cont. Acc.} & \textbf{EM} & \textbf{F1} \\
\midrule
Generic Few-shot & 0.485 & 0.485 & 0.584 \\
Reasoning Few-shot & 0.565 & 0.560 & 0.646 \\
TKG & 0.495 & 0.495 & 0.571 \\
RASTeR & \textbf{0.650} & \textbf{0.640} & \textbf{0.714} \\
\bottomrule
\end{tabular}
\caption{TempReason Needle-in-a-haystack results using 40 semantically similar distractor contexts (based on cosine similarity via \texttt{all-MiniLM-L6-v2}).}
\label{tab:sem_sim_haystack}
\end{table}

RASTeR again achieves the highest performance across all metrics, indicating that it maintains robustness even when distractors are not random but contextually plausible. These findings strengthen the evidence that RASTeR’s reasoning and verification mechanisms generalize beyond artificial perturbations, capturing robustness to realistic retrieval noise.

\subsection{Error Analysis}
\label{appendix:Error Analysis}
Table~\ref{tab:relevance_pred} presents a confusion matrix showing how our relevance reasoner classified different types of context. Note that the pipeline treats both no context and irrelevant context as equivalent, so both the relevance reasoner \textit{should} label them as \textbf{NO / IRR}. As discussed in the main paper, the greatest area for improvement lies in detecting slightly altered contexts, which are only correctly identified 77.1\% of the time.

\label{appendix:ablation}
We evaluate the contribution of individual components in our agentic system by systematically removing steps and comparing the performance of the full system to these ablated variants. As can be seen in Table~\ref{tab:relevance_pred}, Our full model achieves the highest average accuracy across context types, driven by strong performance in the \texttt{none}, \texttt{relevant}, and \texttt{slightly altered} settings. It also maintains a competitive score in the \texttt{relevant} condition, demonstrating balanced robustness across evaluation scenarios.

\begin{table}[t]
\centering
\resizebox{\linewidth}{!}{
\begin{tabular}{lccccc}
\toprule
Ablation& \textbf{none} & \textbf{irrelevant}& \textbf{relevant} & \textbf{slightly altered} & \textbf{avg} \\
\midrule
w/o DateFix         & \textbf{.275}& .263& \textbf{.762}& .200& .375\\
w/o TKG             & \textbf{.275}& .275& .700& .188& .360\\
w/o DetRel          & .150& .212& \textbf{.762}& .175& .325\\
nothing ablated     & .263& \textbf{.300}& .738& \textbf{.250}& .\textbf{388}\\
\bottomrule
\end{tabular}}
\caption{\small Ablation results showing accuracy across different context types. Each row removes a specific module from the full pipeline to assess its contribution.}
\label{tab:ablation}
\end{table}

\label{appendix:Other Metrics}
% EM

% F1

\subsection{Ordering Dates}

\noindent\textbf{Rule order (lexicographic over \textit{key}):}
\begin{enumerate}[itemsep=0pt, topsep=2pt]
  \item Nodes with a \emph{valid starttime} first;
  \item Among equal starttimes: earlier starttime, then higher precision (DAY $<$ MONTH $<$ YEAR $<$ UNKNOWN);
  \item If starttime ties or is invalid: prefer nodes with a \emph{valid endtime}, then earlier endtime, then higher precision;
  \item Nodes with neither valid start nor end appear last; stability preserves their original order.
\end{enumerate}

\noindent\textbf{Note.} Normalized dates (e.g., first day of month/year) are used \emph{only} for ordering and do not overwrite the original strings. Sorting is $O(|S|\log|S|)$; parsing is $O(|S|)$.

\subsection{Prompts}
Both our method and baseline prompts used few-shot examples. To ensure a fair, apples-to-apples comparison, we kept the examples as consistent as possible by using the same set of AlphaGo-related questions and contexts\footnote{We selected the topic AlphaGo randomly; it does not confer any advantage to our method or the baselines and serves solely to ensure consistency across examples.}, randomly selected once and reused throughout. When applicable, we included examples with relevant, irrelevant, slightly-altered, and no context to test model robustness across conditions. Notably, for the Irrelevant Answerer shown in Figure~\ref{fig:no_tkg_reasoning}
, we include only a no-context example, as its pipeline never permits prompting with any other context type. Further details are provided below. 

\subsubsection{RASTeR Prompts}
\label{appendix:prompts:RASTeR}
%%%%%%%%%%%%%%%%%%%%%%%%%%%%%%%%%%%%%%%%%%%%%%%%%%%%%%%%%%%%%%%%%%%%%%%%%%%%%%%%%%%%%%%%%%%%%%%%%%%%%%%%%%%%%%%%%%%%%%%
\vspace{2mm} \noindent \textbf{Relevance Reasoner.}\mbox{}\\
To assess the relevance of a given context, we prompt the model to perform five steps using both the question and the context:  (\textbf{1}) Identify the main entity in the question;  (\textbf{2}) Determine whether this entity appears in the context;  (\textbf{3}) If the context uses pronouns instead of explicit names (e.g., he/she/they instead of `Abraham Lincoln'), assess whether the pronouns plausibly refer to the identified entity;  (\textbf{4}) Evaluate the temporal validity of any dates in the context across four dimensions;  (\textbf{5}) Based on this evaluation, decide whether date correction is needed.  We included five few shot examples for the relevance reasoner: (\textbf{1}) a typical example; (\textbf{2}) an example with a longer context window; (\textbf{3}) an example with some noisy context, (\textbf{4}); an example with longer context and noise; and (\textbf{5}) a counterfactual example. The exact prompt is shown in Figure~\ref{fig:relevance_prompt}. Additionally, we provide an example input-output pair inf Figure~\ref{fig:relevance_agent_example} to demonstrate how the relevance agent can detect temporal inconsistencies without access to golden answers or oracle supervision.

\begin{figure*}[t]
\footnotesize

\centering
\begin{tcolorbox}[colback=gray!5!white, colframe=gray!75!black, title=System Prompt, width=\textwidth]
You will be given a question and a context. Your job is to carefully evaluate the context using the steps below:

\textbf{1. Question Entity:} Identify the main entity of the question.\\

\textbf{2. Does the Entity appear in the context?} \\

\textbf{3. Do Pronouns Refer to the Entity?} \\
If the main entity does not explicitly appear in the context, but the context contains pronouns (e.g., `he'', she'', it''), can these pronouns be reasonably inferred to refer to the entity in question? \\
\textit{Options:} True / False / None (if not applicable)\\

\textbf{4. Evaluation of Context Dates:}
\begin{itemize}
  \item \textbf{a. Chronological Order:} Are the dates logically ordered, or are they contradictory or impossible to sequence?
  \item \textbf{b. Timeframe Alignment:} Do the dates overlap with or lead into the timeframe asked in the question?
  \item \textbf{c. Realism of Time Span:} Is the total date range plausible for the entity, or is it unrealistically broad?
  \item \textbf{d. Historical Consistency:} Do the dates contradict known facts or include anachronisms or future events?
\end{itemize}

\textbf{5. Date Need Correction:} A boolean indicating whether the dates in the context need correction based on the above evaluation.
\end{tcolorbox}

\vspace{1em}

\begin{tcolorbox}[colback=gray!5!white, colframe=gray!75!black, title=User Prompt, width=\textwidth]
Please evaluate the following context by following these steps:
\begin{enumerate}
  \item Question Entity
  \item Does the Entity appear in the context?
  \item Do the Pronouns Refer to the Entity?
  \item Evaluation of Context Dates
  \item Date Need Correction
\end{enumerate}

\textbf{HERE IS THE QUESTION:}\\
\texttt{\{question\}}

\vspace{0.5em}
\textbf{HERE IS THE CONTEXT:}\\
\texttt{\{context\}}
\end{tcolorbox}
\caption{System and user determining the relevance of a provided context.}
\label{fig:relevance_prompt}
\end{figure*}

\vspace{1em}

\begin{figure*}[t]
\centering

% -------- INPUT --------
\begin{tcolorbox}[colback=gray!5!white, colframe=gray!75!black, title=INPUT, width=\textwidth]
\small
\noindent\textbf{Question:} Who was the first Beatle to die?

\vspace{0.5em}
\noindent\textbf{CONTEXT:}
\begin{itemize}[leftmargin=1.5em]
  \item The Beatles were an English rock band formed in Liverpool in 1960.
  \item The group consisted of John Lennon, Paul McCartney, George Harrison, and Ringo Starr.
  \item John Lennon was tragically shot and killed on December 8, 1980.
  \item George Harrison passed away on November 29, 2001, due to lung cancer.
  \item Ringo Starr, born July 7, 1940, is still alive as of 2025.
  \item Paul McCartney died in a car crash in 1966.
  \item Paul McCartney continued to play a central role in the Beatles' later albums, including \textit{Sgt. Pepper's Lonely Hearts Club Band} (1967), \textit{The White Album} (1968), \textit{Abbey Road} (1969), and \textit{Let It Be} (1970).
\end{itemize}
\end{tcolorbox}

\vspace{1em}

% -------- OUTPUT --------
\begin{tcolorbox}[colback=gray!5!white, colframe=gray!75!black, title=OUTPUT, width=\textwidth]
\small
\begin{description}[leftmargin=1.5em, labelsep=0.5em]
  \item[\textbf{Question entity:}] Paul McCartney
  \item[\textbf{Does entity appear in context?}] Yes
  \item[\textbf{Pronoun refers to the entity?}] Not applicable (the context clearly states "Paul McCartney"; no inferences need to be made about He/She/They)
  \item[\textbf{Date Reasoning Chain}] \
  \begin{itemize}[leftmargin=1.5em] \par
    \item \textbf{D1 -- Chronological order:} The claim that Paul died in 1966 comes before Lennon's death in 1980 and Harrison's in 2001. However, the same context states Paul contributed to albums released after 1966. If he were dead, he could not have helped with these later works. This creates a chronological contradiction.
    \item \textbf{D2 -- Timeframe alignment:} The timeframe spans 1966 through 2001, but Paul's supposed death in 1966 does not align with the Beatles' actual release history from 1967--1970.
    \item \textbf{D3 -- Time span realism:} A death in 1966 is unrealistic because it conflicts with Paul's continued presence in recordings and performances through the late 1960s and beyond.
    \item \textbf{D4 -- Historical consistency:} Lennon's 1980 and Harrison's 2001 death dates are historically consistent, but Paul's supposed 1966 death is inconsistent with the known historical record.
  \end{itemize}
  \item[\textbf{Final decision:}] \textsc{Context Needs Correction}
\end{description}
\end{tcolorbox}

\caption{Example input and output of the relevance agent. The date reasoning chain allows the detection temporal consistency (or inconsistency) without gold answers through logical reasoning.}
\label{fig:relevance_agent_example}
\end{figure*}

%%%%%%%%%%%%%%%%%%%%%%%%%%%%%%%%%%%%%%%%%%%%%%%%%%%%%%%%%%%%%%%%%%%%%%%%%%%%%%%%%%%%%%%%%%%%%%%%%%%%%%%%%%%%%%%%%%%%%%%
\vspace{2mm} \noindent \textbf{TKG Constructor.}\mbox{}\\
To incrementally construct a TKG, we prompt the model with a slice of historical context and all previously constructed nodes. The model is asked to identify new temporal facts from the context slice that are not already present in the prior graph. Then convert those facts into structured TKG nodes. Each output node includes: (\textbf{1}) a supporting quote from the context; (\textbf{2}) subject and object entities; (\textbf{3}) their relation; (\textbf{4}) a start and end time; and (\textbf{5}) a reformatted sentence that is grammatically correct, time-grounded, and follows specific templates. The model is explicitly instructed to infer plausible dates when none are stated, use qualifiers like “around” when necessary, and avoid duplicating existing facts. We included two Few-Shot examples to guide the TKG Constructor: (\textbf{1}) an example with no former TKG; and (\textbf{2}) an example with a starting TKG. The exact prompt is shown in Figure~\ref{fig:tkg_constructor_prompt}. As an aside, our prompt encourages the model to format the starttime and endtime in the ``YYYY-DD-MM" format, but does not require it. This intentionally allows varying date granularity. If the context says “He adopted his first dog in 2014,” the model should not hallucinate month/day; forcing YYYY-MM-DD would fabricate precision that is not in the evidence. The changes in format types does not impact LLMs like it would impact a graph database, as LLMs can easily handle slight differences in format. An analysis of the generated time stamps can be found in Table~\ref{tab:timestamp_distribution}
\begin{table*}[t]
\centering
\small
\setlength{\tabcolsep}{8pt}
\renewcommand{\arraystretch}{1.1}
\begin{tabular}{llccccc}
\toprule
\textbf{Model} & \textbf{Dataset} & \textbf{\% YYYYMMDD} & \textbf{\% YYYY} & \textbf{\% MonthYYYY} & \textbf{\% UNKNOWN} & \textbf{N} \\
\midrule
\multirow{4}{*}{\textbf{gemma}} 
 & MenatQA     & 75.10\% & 9.11\% & 0.56\% & 15.23\% & 4,972 \\
 & TempReason  & 99.99\% & 0.00\% & 0.00\% & 0.01\%  & 8,518 \\
 & TimeQA      & 67.93\% & 9.77\% & 0.57\% & 21.74\% & 54,280 \\
\midrule
\multirow{4}{*}{\textbf{gpt}} 
 & MenatQA     & 92.72\% & 0.00\% & 0.00\% & 7.28\%  & 4,588 \\
 & TempReason  & 99.99\% & 0.00\% & 0.00\% & 0.01\%  & 8,444 \\
 & TimeQA      & 81.70\% & 0.00\% & 0.00\% & 18.30\% & 68,990 \\
\midrule
\multirow{4}{*}{\textbf{llama}} 
 & MenatQA     & 85.06\% & 3.14\% & 0.49\% & 11.31\% & 4,898 \\
 & TempReason  & 98.09\% & 0.00\% & 1.90\% & 0.01\%  & 8,330 \\
 & TimeQA      & 62.88\% & 11.91\% & 1.21\% & 24.00\% & 100,046 \\
\bottomrule
\end{tabular}
\caption{Distribution of temporal expression formats across datasets for each model. Percentages indicate the proportion of timestamps expressed as full dates (YYYYMMDD), years (YYYY), month-year pairs (MonthYYYY), or unknown formats.}
\label{tab:timestamp_distribution}
\end{table*}

\begin{figure*}[t]
\footnotesize

\centering
\begin{tcolorbox}[colback=gray!5!white, colframe=gray!75!black, title=System Prompt, width=\textwidth]

You will be presented with a slice of historical context and a previously constructed temporal knowledge graph (TKG). \\
Your task is to identify \textbf{new temporal facts} from the current context and output them as TKG nodes. \\
\textbf{Do not repeat facts already included in the previous TKG.}

\vspace{0.5em}
\noindent Each node should include the following fields:
\begin{itemize}[leftmargin=1.5em]
  \item \textbf{quote}: a verbatim snippet or sentence from the context that supports the node’s validity
  \item \textbf{e1}: subject entity (e.g., person, organization)
  \item \textbf{e2}: object entity (e.g., location, role, other person)
  \item \textbf{rel}: the relation between e1 and e2
  \item \textbf{starttime}: when the relation began
  \item \textbf{endtime}: when the relation ended
  \item \textbf{reformatted}: a rewritten sentence that:
  \begin{itemize}
    \item rearranges the quote to follow the order: time(s), e1, rel, e2
    \item is grammatically correct and includes only e1, e2, rel, and times
    \item \textbf{must include temporal information} (date, year, or month); if not explicit, infer it
    \item uses qualifiers like `around'' or roughly'' when inferring time
    \item follows these example templates:
      \begin{itemize}
        \item \texttt{On \{starttime\}, \{e1\} was \{rel\} \{e2\}}
        \item \texttt{Between \{starttime\} and \{endtime\}, \{e1\} was \{rel\} \{e2\}}
        \item \texttt{Roughly in \{starttime\}, \{e1\} was \{rel\} \{e2\}}
      \end{itemize}
  \end{itemize}
\end{itemize}

\vspace{0.5em}
\noindent Format your output as a list of dictionaries:
\begin{quote}
\texttt{[ \{ "quote": "...", "e1": "...", "e2": "...", "relation": "...", "starttime": "...", "endtime": "...", "reformatted": "..." \}, ... ]}
\end{quote}

\vspace{0.5em}
\noindent\textbf{Notes:}
\begin{itemize}[leftmargin=1.5em]
  \item Begin with an empty TKG or `NONE'' on the first slice.
  \item Only include new nodes clearly grounded in the current context.
  \item Use short, direct quotes.
  \item Do not repeat nodes from the former TKG.
  \item Preferred time formats: `YYYY-MM-DD'', then YYYY'', Month YYYY'', or UNKNOWN''.
  \item You may extract overlapping or nested events if they are distinct.
  \item Use only double quotes in your answer (no single quotes).
\end{itemize}
\end{tcolorbox}

\vspace{1em}

\begin{tcolorbox}[colback=gray!5!white, colframe=gray!75!black, title=User Prompt Header, width=\textwidth]

Construct new TKG nodes using the provided \textbf{context} and \textbf{former\_tkg}.\\
Avoid duplicating facts already extracted. Output only new nodes relevant to the current slice of context.

\vspace{0.5em}
\noindent\textbf{Reminder:} The \texttt{"reformatted"} quote should be a grammatically correct sentence that includes a specific date, year, month, or timespan. If not explicitly stated in the context, \textit{infer it using surrounding information}. In such cases, use terms like `around'' or roughly''.

\vspace{0.5em}
\noindent\textbf{HERE IS THE CONTEXT:}\\
\texttt{\{context\}}

\vspace{0.5em}
\noindent\textbf{HERE IS THE FORMER TKG:}\\
\texttt{\{former\_tkg\}}

\end{tcolorbox}
\caption{System and user prompt for generating new temporal knowledge graph (TKG) slices of a provided context.}
\label{fig:tkg_constructor_prompt}
\end{figure*}
%%%%%%%%%%%%%%%%%%%%%%%%%%%%%%%%%%%%%%%%%%%%%%%%%%%%%%%%%%%%%%%%%%%%%%%%%%%%%%%%%%%%%%%%%%%%%%%%%%%%%%%%%%%%%%%%%%%%%%%

\vspace{2mm} \noindent \textbf{TKG Date Completion.}\mbox{}\\
When the Relevance Reasoner determines that a context requires correction, we manually remove the \texttt{starttime} and \texttt{endtime} from each node in the TKG, replacing them with placeholder values $X$ and $Y$. The model is then prompted to (\textbf{1}) infer plausible temporal bounds using historical knowledge or contextual cues, and (\textbf{2}) generate a natural, grammatically correct sentence that incorporates the subject (\texttt{e1}), relation (\texttt{rel}), object (\texttt{e2}), and the inferred timeframe. The output must include both the completed sentence and the recovered temporal fields in a structured format. This step allows the model to use its internal knowledge to infer temporal boundaries, enabling accurate correction of incomplete TKG facts. We included a single few shot example to guide the model. The exact prompt is shown in Figure~\ref{fig:tkg_corrector_prompt}.

\begin{figure*}[t]
\footnotesize

\centering
\begin{tcolorbox}[colback=gray!5!white, colframe=gray!75!black, title=System Prompt, width=\textwidth]

You are given a temporal knowledge graph (TKG) triple with missing \texttt{starttime} and \texttt{endtime}.

\vspace{0.5em}
\noindent\textbf{Your task is to:}
\begin{enumerate}
    \item Infer appropriate \texttt{starttime} and \texttt{endtime} based on historical knowledge or reasonable assumptions.
    \item Write a grammatically correct and natural-sounding sentence that incorporates:
    \begin{itemize}
        \item the subject (\texttt{e1})
        \item the relationship (\texttt{rel})
        \item the object (\texttt{e2})
        \item and the inferred temporal range
    \end{itemize}
\end{enumerate}

You are allowed to rephrase the sentence as long as all elements are included and the timeframe is clearly conveyed.

\vspace{0.5em}
\noindent\textbf{Your output should follow this format:}
\begin{quote}
\texttt{COMPLETE SENTENCE: [your complete sentence]} \\
\texttt{STARTIME: [YYYY-MM-DD]} \\
\texttt{ENDTIME: [YYYY-MM-DD]}
\end{quote}

\vspace{0.5em}
\noindent\textbf{Example:}
\begin{quote}
\textbf{INPUT:} \\
\texttt{\{"e1": "Arseny Dmitrievich Mironov", "e2": "USSR State Prize", "rel": "recipient of", "starttime": X, "endtime": Y\}}

\vspace{0.5em}
\textbf{OUTPUT:} \\
\texttt{COMPLETE SENTENCE: Arseny Dmitrievich Mironov received the USSR State Prize in 1976.} \\
\texttt{STARTIME: 1976-01-01} \\
\texttt{ENDTIME: 1976-12-31}
\end{quote}
\end{tcolorbox}

\vspace{1em}

\begin{tcolorbox}[colback=gray!5!white, colframe=gray!75!black, title=User Prompt Header, width=\textwidth]

Given the following TKG triple with missing temporal information, fill in the \texttt{starttime} and \texttt{endtime}, and write a complete, natural-sounding sentence.

\vspace{0.5em}
\noindent\textbf{You must:}
\begin{itemize}
    \item Include \texttt{e1}, \texttt{rel}, and \texttt{e2}
    \item Clearly indicate the time period
    \item Use correct grammar and phrasing
\end{itemize}

\vspace{0.5em}
\noindent\textbf{Format:}
\begin{quote}
\texttt{COMPLETE SENTENCE: ...} \\
\texttt{STARTIME: ...} \\
\texttt{ENDTIME: ...}
\end{quote}

\vspace{0.5em}
\noindent\textbf{HERE IS YOUR TRIPLE:} \\
\texttt{\{triple\}}

\end{tcolorbox}
\caption{System and user prompt for inferring missing temporal values in a temporal knowledge graph triple.}
\label{fig:tkg_corrector_prompt}
\end{figure*}

\vspace{2mm} \noindent \textbf{Relevant Answerer.}\mbox{}\\
After the construction of the TKG, we prompt the model with a question and the TKG. The model is instructed to perform three steps: (\textbf{1}) select the node(s) from the TKG that are temporally relevant and contain information necessary to answer the question; (\textbf{2}) explain how the selected node(s) support the answer, including reasoning over temporal relationships such as before/after conditions; and (\textbf{3}) provide a confident, direct answer based on the evidence, or make an educated guess using indirect cues if no explicit answer is available. This step leverages the structured context encoded in the TKG to produce grounded, time-aware answers. We included six Few-Shot examples to help guide the models reasoning: (\textbf{1}) an example with relevant context; (\textbf{2}) and example without context; (\textbf{3}) an example with random context; (\textbf{4}) and example with slightly altered context; (\textbf{5}) an example showing when pronouns correctly refer to the entity in the question; and (\textbf{6}) an example showing when the pronouns do not refer to the entity in the question. The exact prompt is shown in Figure~\ref{fig:tkg_reasoning}.

\begin{figure*}[t]
\centering
\footnotesize

\begin{tcolorbox}[colback=gray!5!white, colframe=gray!75!black, title=System Prompt, width=\textwidth]

You are given a question and a temporal knowledge graph (TKG). Your job is to answer the question using the TKG to assist you.

\vspace{0.5em}
\noindent\textbf{Please follow these steps:}

\vspace{0.25em}
\noindent\textbf{1. Select Supporting Nodes:}
\begin{itemize}[leftmargin=1.5em]
  \item From the TKG, return the node(s) that provide the information necessary to answer the question.
  \item You may include one or more nodes.
  \item Only include nodes that are temporally relevant to the question.
  \item You must consider the time frame mentioned in the question.
  \item If multiple matching nodes exist, include them all.
\end{itemize}

\vspace{0.25em}
\noindent\textbf{2. Explain Your Reasoning:}
\begin{itemize}[leftmargin=1.5em]
  \item Justify how the node(s) support your answer.
  \item If no node is directly about the question, you may infer the answer from strong contextual clues.
  \item For \textit{before}/\textit{after} questions, identify the event that occurred immediately before or after the referenced one.
  \item \textbf{Example:} \\
  \textit{Context:} Dan attended high school from 2010–2014, undergrad from 2014–2018, a master's from 2023–2024, and began a PhD in 2024.\\
  \textit{Question:} What did Dan do after high school?\\
  \textit{Reasoning:} Dan completed undergrad, a master's, and began a PhD after high school. However, undergrad was immediately after, so it is the correct answer.
\end{itemize}

\vspace{0.25em}
\noindent\textbf{3. Answer the Question:}
\begin{itemize}[leftmargin=1.5em]
  \item Respond in the format: \texttt{The answer is X}
  \item If no nodes provide a direct answer, use indirect evidence to make an educated guess.
  \item For instance, political roles, awards, institutions, or cities may imply nationality or affiliation.
  \item Your answer should be confident and definite.
\end{itemize}

\vspace{0.25em}
\noindent\textbf{Note:} Use only double quotes in your answer. Do not use single quotes.

\end{tcolorbox}

\vspace{1em}

\begin{tcolorbox}[colback=gray!5!white, colframe=gray!75!black, title=User Prompt Header, width=\textwidth]

Given a question and a temporal knowledge graph (TKG), answer the question using the TKG to assist you.

\vspace{0.25em}
\noindent\textbf{Follow these steps:}
\begin{enumerate}
  \item Select Supporting Nodes
  \item Explain Your Reasoning
  \item Answer the Question
\end{enumerate}

If no nodes directly provide information to answer the question, use indirect evidence to make an educated guess.

\vspace{0.5em}
\noindent\textbf{HERE IS YOUR QUESTION:}\\
\texttt{\{question\}}

\vspace{0.5em}
\noindent\textbf{HERE IS THE TKG:}\\
\texttt{\{TKG\}}

\end{tcolorbox}
\caption{System and user prompts for answering temporal questions using a temporal knowledge graph (TKG).}
\label{fig:tkg_reasoning}
\end{figure*}

\vspace{2mm} \noindent \textbf{Irrelevant Answerer.}\mbox{}\\
When the context is determined to be irrelevant, we discard the context, and prompt the model to answer questions using only its internal knowledge. The model is guided through a 3-step reasoning process: (\textbf{1}) restate the question to clarify what is being asked; (\textbf{2}) reason toward an answer using general world knowledge; and (\textbf{3}) provide a final answer in a clear, structured format. By discarding the context, we eliminate distractors and evaluate the model’s ability to interpret and answer temporal questions without relying on a retrieved context. We provide a single Few-Shot example to help guide the model's reasoning. The exact prompt is shown in Figure~\ref{fig:no_tkg_reasoning}.

\begin{figure*}[t]
\footnotesize

\centering
\begin{tcolorbox}[colback=gray!5!white, colframe=gray!75!black, title=System Prompt, width=\textwidth]

You are a helpful and precise assistant. When given a question, respond clearly and concisely, only using relevant information. To help you arrive at the correct answer, follow this 3-step reasoning process:

\vspace{0.5em}
\noindent\textbf{1. Restate the Question} \\
Rephrase the original question to clarify what is being asked.

\vspace{0.5em}
\noindent\textbf{2. Reason} \\
Rely on your general knowledge to reason toward an answer.

\vspace{0.5em}
\noindent\textbf{3. Answer} \\
Provide your final answer using the format: \texttt{The answer is X.}

\vspace{0.5em}
\noindent\textbf{Note:} Only use double quotes in your answer. Do not use single quotes.

\end{tcolorbox}

\vspace{1em}

\begin{tcolorbox}[colback=gray!5!white, colframe=gray!75!black, title=User Prompt Header, width=\textwidth]

Please answer the question using the provided context. \\
Follow the 3-step reasoning process above, and end with a final answer in Step 3.

\vspace{0.5em}
\noindent\textbf{HERE IS THE QUESTION:}\\
\texttt{\{question\}}

\end{tcolorbox}
\caption{System and user prompts for answering questions without context using a structured 3-step reasoning process.}
\label{fig:no_tkg_reasoning}
\end{figure*}

\label{appendix:prompts:baselines}
\subsubsection{Baseline Prompts}
All baseline prompts have four Few-Shot examples to help guide their reasoning: (\textbf{1}) an example with relevant context; (\textbf{2}) an example without context; (\textbf{3}) an example with random context; (\textbf{4}) an example with slightly altered context.

\vspace{2mm} \noindent \textbf{Few-Shot.}\mbox{}\\
Our first baseline evaluates model performance using a minimal prompt that mirrors common few-shot setups. The model is given a question and a corresponding context and is instructed to respond concisely using the format: \texttt{The answer is X}. Unlike our structured approaches, this prompt includes no explicit reasoning steps or guidance for interpreting the context. It serves to benchmark how well the model can extract answers when given relevant input, and how it performs in the presence of no or irrelevant context without any reasoning scaffolding. The exact prompt is shown in Figure~\ref{fig:few-shot_prompt}.

\begin{figure*}[t]
\footnotesize

\centering
\begin{tcolorbox}[colback=gray!5!white, colframe=gray!75!black, title=System Prompt, width=\textwidth]

You are a helpful and precise assistant. When given a context and a question, respond clearly and concisely, only using relevant information.

\vspace{0.5em}
\noindent Please use the format: \texttt{The answer is X} where \texttt{X} is your answer.

\vspace{0.5em}
\noindent\textbf{Example formatting:}
\begin{quote}
\texttt{The answer is Michael Phelps.}
\end{quote}

\end{tcolorbox}

\vspace{1em}

\begin{tcolorbox}[colback=gray!5!white, colframe=gray!75!black, title=User Prompt Header, width=\textwidth]

Please answer the question given the context. \\
Your response should follow the format: \texttt{The answer is X}

\vspace{0.5em}
\noindent\textbf{HERE IS THE QUESTION:}\\
\texttt{\{question\}}

\vspace{0.5em}
\noindent\textbf{HERE IS THE CONTEXT:}\\
\texttt{\{context\}}

\end{tcolorbox}
\caption{System and user prompts for answering questions with concise, format-specific responses.}
\label{fig:few-shot_prompt}
\end{figure*}

\vspace{2mm} \noindent \textbf{Reasoning.}\mbox{}\\
Our second baseline introduces a structured 5-step reasoning process to guide the model through question answering. Given a question and a context, the model is instructed to (\textbf{1}) restate the question to clarify its intent; (\textbf{2}) assess whether the context is relevant; (\textbf{3}) quote specific evidence from the context, or indicate \texttt{NONE} if no useful information is found; (\textbf{4}) reason toward an answer using either the provided evidence or its own internal knowledge; and (\textbf{5}) produce a final answer in the format: \texttt{The answer is X}. This format encourages explicit reasoning and evidence grounding. The exact prompt is shown in Figure~\ref{fig:Reasoning}.

\begin{figure*}[t]
\footnotesize

\centering
\begin{tcolorbox}[colback=gray!5!white, colframe=gray!75!black, title=System Prompt, width=\textwidth]

You are a helpful and precise assistant. When given a context and a question, respond clearly and concisely, only using relevant information. To help you arrive at the correct answer, follow this 5-step reasoning process:

\vspace{0.5em}
\noindent\textbf{1. Restate the Question} \\
Rephrase the original question to clarify what is being asked.

\vspace{0.5em}
\noindent\textbf{2. Evaluate Context Relevance} \\
Determine whether the provided context contains information that is useful for answering the question.

\vspace{0.5em}
\noindent\textbf{3. Quote Supporting Evidence} \\
Copy and paste the exact portion(s) of the context that support your answer. \\
If no useful evidence exists, write \texttt{NONE}.

\vspace{0.5em}
\noindent\textbf{4. Reason} \\
If the context is relevant, use the quoted evidence to logically derive the answer. \\
If the context is not relevant, rely on your general knowledge to reason toward an answer.

\vspace{0.5em}
\noindent\textbf{5. Answer} \\
Provide your final answer using the format: \texttt{The answer is X.}

\end{tcolorbox}

\vspace{1em}

\begin{tcolorbox}[colback=gray!5!white, colframe=gray!75!black, title=User Prompt Header, width=\textwidth]

Please answer the question using the provided context. \\
Follow the 5-step reasoning process above, and end with a final answer in Step 5.

\vspace{0.5em}
\noindent\textbf{HERE IS THE QUESTION:} \\
\texttt{\{question\}}

\vspace{0.5em}
\noindent\textbf{HERE IS THE CONTEXT:} \\
\texttt{\{context\}}

\end{tcolorbox}
\caption{System and user prompts for answering questions using a structured 5-step reasoning process.}
\label{fig:Reasoning}
\end{figure*}

\vspace{2mm} \noindent \textbf{Simple TKG.}\mbox{}\\
This baseline a non-iterative TKG construction without the full multi-agent pipeline. The model is prompted to (\textbf{1}) extract all entities from the context, including people, places, roles, and other named concepts; (\textbf{2}) construct a TKG; and (\textbf{3}) answer the question based on the constructed TKG using the standard format: \texttt{The answer is X}. The model is allowed to correct factual inconsistencies in the context or fall back on internal knowledge when context is irrelevant. This prompt provides a basic measure of how well the model can extract temporal structure and reason over it in a single pass. The exact prompt is shown in Figure~\ref{fig:simple_tkg}.

\begin{figure*}[t]
\footnotesize
\centering
\begin{tcolorbox}[colback=gray!5!white, colframe=gray!75!black, title=System Prompt, width=\textwidth]

You are a helpful and precise assistant. You will be presented with some context and a question. Your job has two parts.

\vspace{0.5em}
\noindent\textbf{First: Identify all entities} in the context, including places, names, occupations, and things. \\
List them using the following format, wrapped in triple backticks. Do not skip any entities.
\begin{quote}
\texttt{
} \\
\texttt{e1. Yoko Ono} \\
\texttt{e2. Businessman} \\
\texttt{e3. Europe} \\
\texttt{
}
\end{quote}

\vspace{0.5em}
\noindent\textbf{Second: Construct a Temporal Knowledge Graph (TKG)} based on the context using the identified entities. \\
Each TKG node should include the following fields:
\begin{itemize}[leftmargin=1.5em]
  \item \texttt{Entity1}
  \item \texttt{Entity2}
  \item \texttt{Relation}
  \item \texttt{Timestamp}
\end{itemize}

\noindent\textbf{Format:}
\begin{quote}
\texttt{[} \\
\quad \texttt{\{'Entity1': '...', 'Entity2': '...', 'Relation': '...', 'Timestamp': '...'\},} \\
\quad \texttt{...} \\
\texttt{]}
\end{quote}

\vspace{0.5em}
\noindent\textbf{Additional Instructions:}
\begin{itemize}[leftmargin=1.5em]
  \item If the context is partially incorrect, correct the information before building that part of the TKG.
  \item If the context is irrelevant or marked as \texttt{NONE}, discard it and use your internal knowledge instead.
\end{itemize}

\vspace{0.5em}
\noindent Once the TKG is complete, use it to answer the question. Respond concisely using the format: \texttt{The answer is X.}

\vspace{0.5em}
\noindent\textbf{Example formatting:}
\begin{quote}
\texttt{The answer is Michael Phelps.}
\end{quote}

\end{tcolorbox}

\vspace{1em}

\begin{tcolorbox}[colback=gray!5!white, colframe=gray!75!black, title=User Prompt Header, width=\textwidth]

Build a temporal knowledge graph (TKG) to help answer the question using the provided context. \\
The TKG should be a list of nodes, each with \texttt{Entity1}, \texttt{Entity2}, \texttt{Relation}, and \texttt{Timestamp} fields.

\vspace{0.5em}
Once the TKG is complete, use it to answer the question. \\
Your answer should follow the format: \texttt{"The answer is X"}

\vspace{0.5em}
\noindent\textbf{HERE IS THE QUESTION:} \\
\texttt{\{question\}}

\vspace{0.5em}
\noindent\textbf{HERE IS THE CONTEXT:} \\
\texttt{\{context\}}

\end{tcolorbox}
\caption{System and user prompts for entity extraction, temporal knowledge graph construction, and question answering.}
\label{fig:simple_tkg}
\end{figure*}

\end{document}